\pdfoutput=1
\documentclass[11pt]{article}
\usepackage[preprint]{acl}

\usepackage{times}
\usepackage{latexsym}
\usepackage[T1]{fontenc}
\usepackage[utf8]{inputenc}
\usepackage{microtype}
\usepackage{inconsolata}
\usepackage{graphicx}

\usepackage{amsmath}
\usepackage{amssymb}
\usepackage{amsthm}
\usepackage{mathtools}
\usepackage{bm}
\usepackage{booktabs}
\usepackage{array}
\usepackage{colortbl}
\usepackage{placeins}
\usepackage{multirow}
\usepackage{xcolor}
\usepackage{subcaption}
\usepackage{enumitem}

\newcommand{\Nker}{\mathcal{N}_{R}}
\newcommand{\Nvis}{\mathcal{N}_{\mathrm{vis}}}
\newcommand{\Ninv}{\mathcal{N}_{\mathrm{inv}}}
\newcommand{\bk}{b^{\kappa}}
\newcommand{\piref}{\pi_{\mathrm{ref}}}
\newcommand{\pistar}{\pi^{*}_{\beta}}
\newcommand{\KL}{\mathrm{KL}}
\newcommand{\cls}{\mathrm{cls}}
\newcommand{\simR}{\sim_{R}}
\newcommand{\simF}{\sim_{F}}

\newcommand{\simpi}{\sim_{\pi}}

\newcommand{\Exp}{\mathbb{E}}
\newcommand{\one}{\mathbf{1}}

\newcommand{\Rvis}{R_{\mathrm{vis}}}
\newcommand{\Rinv}{R_{\mathrm{inv}}}
\newcommand{\Ufilt}{U_{\mathrm{filt}}}

\newcommand{\erased}{\emph{erased}}
\newcommand{\rerouted}{\emph{rerouted}}
\newcommand{\rescaled}{\emph{rescaled}}

\definecolor{tabzebra}{HTML}{EEF1F5}
\definecolor{tabgroup}{HTML}{D9DFE8}

\newtheorem{proposition}{Proposition}

\title{Erased, Rerouted, or Rescaled? Post-Training and the Causal Quotient of a Language Model's Belief State}

\author{
  \textbf{Weihan Li}\textsuperscript{$\ast$1},
  \textbf{Tianshi Zheng}\textsuperscript{$\ast$2},
  \textbf{Junhao Wu}\textsuperscript{3},
  \textbf{Xinlei Chen}\textsuperscript{4}
  \\
  \textsuperscript{1}The University of Tokyo \quad
  \textsuperscript{2}The Hong Kong University of Science and Technology
  \\
  \textsuperscript{3}RWTH Aachen University \quad
  \textsuperscript{4}Harbin Institute of Technology, Shenzhen
  \\
  \href{mailto:liweihan1107@g.ecc.u-tokyo.ac.jp}{\textcolor{black}{\texttt{liweihan1107@g.ecc.u-tokyo.ac.jp}}}
}

\begin{document}
\maketitle
\begingroup
\renewcommand{\thefootnote}{\fnsymbol{footnote}}
\footnotetext[1]{Equal Contribution}
\endgroup

\begin{abstract}
What happens to information a pretrained model already encodes when post-training no longer rewards using it? The common language of representation \emph{compression} conflates three fates: information may be \emph{erased}, \emph{rerouted} away from the decision while still represented, or \emph{rescaled} to occupy less variance while still represented and used. We make these fates identifiable in models whose pretraining recovers Bayesian belief states. A reward that reads only a coarse function of the hidden state defines an exact reward-null kernel. The kernel lets us separately measure whether the information remains recoverable, whether decisions causally depend on it, and how much activation variance it occupies. Theory says what is protected: KL-anchored reinforcement learning preserves the reference policy's log-odds among equally rewarded outputs, supervised and unanchored objectives carry no such constraint, and spectral compression implies neither erasure nor loss of use. In controlled worlds, post-training mostly reroutes or rescales reward-null information and leaves it decodable. Without an anchor decisions can stop using it although the representation survives, and with one they keep using it. Erasure appears only under prolonged weight decay, for distinctions that neither reward nor next-token prediction can see. Open language models show the same dissociation: in-context belief geometry stays decodable under late-layer spectral compression, and within-class behavior depends on the anchor. Post-training thus selects a causal quotient of the pretrained belief state: the reward defines decision-equivalence, the anchor and the state update protect part of what it ignores, and optimization decides whether the rest is erased, rerouted, or rescaled.
\end{abstract}

\section{Introduction}
\label{sec:intro}

\begin{figure*}[t]
  \centering
  \includegraphics[width=\textwidth]{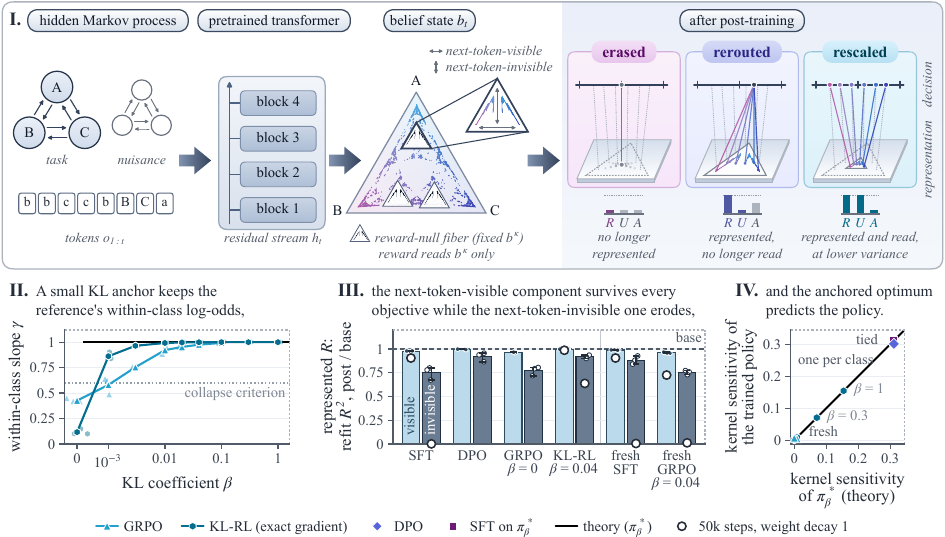}
  \caption{\textbf{From a belief state to its fates.} \textbf{I.} A hidden Markov process with a task and a nuisance variable generates tokens, and a pretrained transformer represents the belief state $b_t$. The reward reads only the class marginal $\bk$ (large simplex). Beliefs that share it form a reward-null fiber (small simplices), with one direction the next token can see and one it cannot. Each card shows one fiber in the representation (plane) and the decisions its points map to (track; dashed: the base model): \erased{} (collapsed), \rerouted{} (intact, every point mapped to one decision) or \rescaled{} (shrunk, mapped to the same decisions as before), with the fate's signature in $R$, $U$ and $A$. \textbf{II--IV.} Three results in preview (open circles in III: long training with weight decay).}
  \label{fig:overview}
\end{figure*}

When a pretrained transformer is post-trained to act rather than to predict, what happens to distinctions it has already learned but the new objective no longer rewards using? A predictor must keep every distinction among pasts that changes what comes next, which for hidden-Markov sequences is the \emph{belief state} \citep{crutchfield1989,shalizi2001}, the Bayesian posterior over hidden states. A downstream reward that reads only a coarse function of the hidden state asks for less. It induces a quotient of the belief state, in which predictively distinct histories become decision-equivalent, and task performance leaves the fate of the distinctions inside each class open. They may be \erased{}, \rerouted{} away from the decision while remaining represented, or \rescaled{} to occupy less variance while remaining represented and used (Figure~\ref{fig:overview}).

The belief state is by now a measurable object. Models trained on hidden-Markov sequences represent it linearly in their residual streams \citep{shai2024,piotrowski2025}, and open language models reconstruct the same geometry in context and use it causally \citep{balcells2026}. Crucially, these models also retain belief components that the next token cannot see: a structured state can be decoded and intervened on even where an objective does not read it.

Studies of post-training observe representational change without knowing what should remain relevant: rewritten or reweighted behavior-relevant geometry \citep{wang2026}, spectra that shift across training stages \citep{li2025}, altered reasoning trajectories \citep{sun2026}, and fine-tuning that wraps existing circuits \citep{jain2024,prakash2024,lee2024}. Such measurements cannot say what kind of information changed: a shrinking spectrum can accompany information loss, a change in which directions the policy reads, or an invertible rescaling that leaves information and behavior intact.

We resolve this ambiguity in hidden Markov worlds where decision-equivalence is known before post-training begins. The hidden state pairs a task variable with a nuisance variable, the belief state is known exactly, and a reward that reads only the task variable defines in closed form the \emph{reward-null kernel}: distinctions that pretraining has reason to encode and reward has no reason to read. Theory says which of these distinctions are protected before any network is trained. The optimum of KL-regularized reinforcement learning is $\pistar\propto\piref\,e^{q/\beta}$ \citep{korbak2022,rafailov2023,gxchen2025}, so equally rewarded actions keep exactly the reference policy's log-odds at every $\beta$, and where these depend on a reward-null direction the anchored policy must stay sensitive to it. Supervised and unanchored objectives carry no such constraint, and spectral statistics are invariants of neither information nor its use. We therefore ask three separate questions of each reward-null distinction: is it still \emph{represented}, does the policy still \emph{use} it, and how much variance is \emph{allocated} to it?

The resulting picture is sharper than a generic account of post-training as compression. Reward-null information is usually retained after its role in the policy changes: what next-token prediction can see stays decodable under all practical objectives while its variance contracts, and the anchor decides whether the decision keeps reading it. Genuine erasure appears only under persistent pressure such as prolonged weight decay, for distinctions protected by neither reward nor prediction. A recurrent filter can keep encoding a distinction it has stopped using. What the reward leaves open, optimization dynamics resolve: without an anchor, policies settle on within-class behavior far from the reference model's, all of it equally rewarded.

The same dissociation appears in open language models. Post-training leaves the in-context belief geometry, including components invisible to the next token, decodable while late-layer spectra contract, and under coarse-class GRPO within-class preferences leave the base policy without the anchor and return to it with one. Post-training thus selects a \emph{causal quotient} of the belief state, the distinctions its decisions still act on, and erased, rerouted, and rescaled are distinct, empirically identifiable fates of reward-null information.
\section{Quotients of a Belief State}
\label{sec:setting}

\paragraph{Belief states.}
A hidden Markov model with hidden states $s$, transition matrix $T$ and emission matrix $E$ generates tokens $o_1,o_2,\dots$ The belief state after a history $h=o_{1:t}$ is the posterior over the state that emits the next token, $b(h)=P(s_{t+1}\mid o_{1:t})$. It is updated by Bayesian filtering, the next-token distribution is $b(h)^{\top}E$, and two histories with the same belief share every future. Networks pretrained on such sequences represent $b$ linearly in their activations \citep{shai2024}, and so do our base models (Appendix~\ref{app:setup}).

\paragraph{The reward quotient.}
Hidden states are pairs $s=(c,f)$ of a task variable and a nuisance variable. At a query token, which also occurs at random positions during pretraining, the model chooses an action $a$ and receives $r(s,a)=\one[\cls(a)=c]$. The expected reward $q(b)=Mb$ is linear in the belief and depends on it only through the class marginal $\bk$. Two beliefs are \emph{reward-equivalent}, $b\simR b'$, when $q(b)=q(b')$, that is, when $b-b'$ lies in the \emph{reward-null kernel}
\begin{equation}
\Nker=\ker M\cap\one^{\perp},
\label{eq:kernel}
\end{equation}
which is computed from the parameters of the world before any network is trained. Each direction of $\Nker$ is a distinction among pasts that prediction needs and the reward never reads, and the beliefs sharing one $\bk$ form a \emph{fiber}. The next token divides the kernel further. The \emph{next-token-invisible component} $\Ninv=\{d\in\Nker:\ d^{\top}E=0\}$ changes neither the reward nor the next-token distribution and matters only for later tokens, the kind of component \citet{balcells2026} found language models to maintain. Its orthogonal complement within the kernel, $\Nvis=\Nker\cap\Ninv^{\perp}$, is the \emph{next-token-visible component}. We shorten the two names to invisible and visible component once the context has fixed them.

\paragraph{The causal quotient.}
The reward quotient records what the reward cannot see. What a policy acts on is a second partition. For a policy $\pi$ let $h\simpi h'$ when $\pi(\cdot\mid h)=\pi(\cdot\mid h')$. We call the partition of histories induced by $\simpi$ the \emph{causal quotient} of $\pi$, because membership is tested by intervening on the history and observing the decision, which decoding from the activations cannot replace. The reference policy of a pretrained model separates histories that the reward does not, and post-training selects a new causal quotient: the objective constrains it (Section~\ref{sec:theory}) and optimization chooses within the constraints (Section~\ref{sec:direction}).

\begin{table}[t]
\centering
\small
\setlength{\tabcolsep}{4pt}
\renewcommand{\arraystretch}{1.12}
\begin{tabular}{c>{\raggedright\arraybackslash}p{0.34\columnwidth}>{\raggedright\arraybackslash}p{0.46\columnwidth}}
\toprule
 & Question about $S$ & Operational meaning \\
\midrule
\rowcolor{tabzebra} $R_{S}$ & Is it still represented? & linear recoverability of the belief's coordinates along $S$ \\
$U_{S}$ & Does the decision still use it? & causal sensitivity of the policy to $S$ \\
\rowcolor{tabzebra} $A_{S}$ & How much space does it occupy? & share of the activation variance carried by $S$ \\
\bottomrule
\end{tabular}
\caption{Three coordinates of a reward-null subspace $S$.}
\label{tab:rua}
\end{table}

\paragraph{Three questions about a reward-null distinction.}
For a subspace $S$ of the kernel, the whole of it or one component, we ask the three questions of Table~\ref{tab:rua} of a post-trained network and report each answer as a ratio to the base model. The worlds supply the intervention behind $U$. Two histories that share their task symbols and differ in their nuisance symbols have identical $\bk$ and differ only along the kernel, so the response of the policy to exchanging such \emph{twins} is the causal effect of the kernel on the decision, measured without a probe. We report $R$ for each component ($\Rvis$, $\Rinv$) and, unless a subscript says otherwise, $U$ and $A$ for the whole kernel, along which twins differ. The estimators build on usable information for probing \citep{xu2020}, interchange interventions \citep{geiger2021} and effective-rank statistics \citep{garrido2023} (Appendix~\ref{app:measure}). A distinction is \erased{} when $R$ falls, \rerouted{} when $R$ stays and $U$ falls, so that it leaves the causal quotient while remaining in the representation, and \rescaled{} when $R$ and $U$ stay and $A$ falls.

\paragraph{Allocation is a coordinate of its own.}
\begin{proposition}[Spectral statistics are not invariants]
\label{prop:gauge}
Let a network factor as $y=g(x)$ through the activations $x$ of some layer, and let $G$ be an invertible matrix. The network $(Gx,\ g\circ G^{-1})$ computes the same function, gives the same value to every probe-based measure of information whose probe family is closed under $G$, and responds identically to every interchange intervention, while its activation covariance $G\Sigma G^{\top}$ can be any positive semidefinite matrix of the rank of $\Sigma$.
\end{proposition}

Effective rank, spectral decay and the variance share of any subspace can therefore take any feasible value while represented and used information stay fixed. \textbf{Spectral compression alone cannot identify information loss.} Within one architecture, whose normalization layers leave only rotations as exact symmetries, allocation is still a well-defined property of the trained solution, and it has to be measured separately from the other two.
\section{Which Reward-Null Distinctions Are Protected}
\label{sec:theory}

That the reward ignores a distinction does not make it dispensable. The reward defines which distinctions are decision-equivalent. Two constraints, both computable before training, then determine which reward-null distinctions remain protected: the anchor, through the reference policy, and the state update, through the filter. Optimization decides the fate of the rest.

\paragraph{The anchor preserves the reference's distinctions.}
KL-regularized reinforcement learning maximizes the expected reward minus $\beta\,\KL(\pi\,\|\,\piref)$, and its optimum tilts the reference by the expected reward $q$,
\begin{equation}
\pistar(a\mid h)\;\propto\;\piref(a\mid h)\,\exp\!\big(q(h)_{a}/\beta\big).
\label{eq:tilt}
\end{equation}

\begin{proposition}[Anchor identity]
\label{prop:anchor}
For every $\beta>0$: \emph{(i)} if two actions have equal expected reward at $h$, their log-odds under $\pistar$ equal their log-odds under $\piref$; \emph{(ii)} if $b(h)\simR b(h')$, the difference between the two histories in the log-odds of any pair of actions is the same under $\pistar$ as under $\piref$.
\end{proposition}

Both statements hold because the tilt depends on the history only through $q$, and (i) is known at the level of the optimum \citep{korbak2022,rafailov2023,gxchen2025}. What we use is its consequence for the belief state. Many outputs of a language model are equally correct, the reward cannot separate them, and the reference is the model's own next-token distribution. Our main design reproduces this situation: every token is an action, the reward reads only the token's class, and the reference prefers among the tied tokens of a class according to the nuisance belief. The optimal policy then depends on the belief only through $(\bk,\,b^{\top}E)$. Its causal quotient merges exactly the beliefs that differ along $\Ninv$, at every $\beta$, and $\Nvis$ must stay represented and used at the decision. In centered log-probabilities the optimum is $C\log\piref+Cq/\beta$, so the variation within a fiber is untouched while the variation between fibers grows as $1/\beta$. The kernel keeps its log-odds and loses its share of the variance: at the output, anchored reinforcement learning is rescaling by construction.

\begin{table}[t]
\centering
\footnotesize
\setlength{\tabcolsep}{3pt}
\renewcommand{\arraystretch}{1.1}
\begin{tabular}{>{\raggedright\arraybackslash}p{0.26\columnwidth}>{\raggedright\arraybackslash}p{0.66\columnwidth}}
\toprule
Variant & What it isolates \\
\midrule
\rowcolor{tabgroup}\multicolumn{2}{l}{\textit{Worlds}}\\
\rowcolor{tabzebra} separable & independent task and nuisance chains: the filter needs none of the kernel, and half of it is next-token-invisible \\
visible control & the same with the whole kernel next-token-visible \\
\rowcolor{tabzebra} entangled & filtering the task variable requires the nuisance belief \\
Mess3 & the process of \citet{shai2024} \\
\specialrule{0.3pt}{0pt}{0pt}
\rowcolor{tabgroup}\multicolumn{2}{l}{\textit{Action designs}}\\
\rowcolor{tabzebra} tied & every token is an action and the reward reads its class: the tokens of a class are tied, and the reference prefers among them \\
one per class & no ties: Proposition~\ref{prop:anchor} predicts the policy's sensitivity \\
\rowcolor{tabzebra} fresh tokens & a uniform reference, the built-in control \\
\bottomrule
\end{tabular}
\caption{Experiment map. Objectives: SFT on a demonstrator's labels, DPO, GRPO, REINFORCE and KL-regularized reinforcement learning with the exact gradient (KL-RL), over anchor strengths $\beta$ that include 0.}
\label{tab:map}
\end{table}

\paragraph{Supervision and unanchored reward protect nothing at the decision.}
A demonstrator who knows the class, and whose choice among equally correct tokens ignores the nuisance of the history, gives supervised fine-tuning the optimum $\pi(a\mid b)=\sum_{k}\bk_{k}\,p_{\mathrm{demo}}(a\mid k)$. It factors through the reward quotient, and the whole kernel is free at the decision. The contrast with the anchored optimum comes from who chooses among equally rewarded answers: the model itself, or the author of the data. Without the anchor every policy supported on the best class is optimal, and the objective is silent about behavior within a class.

\begin{figure*}[t]
  \centering
  \includegraphics[width=\textwidth]{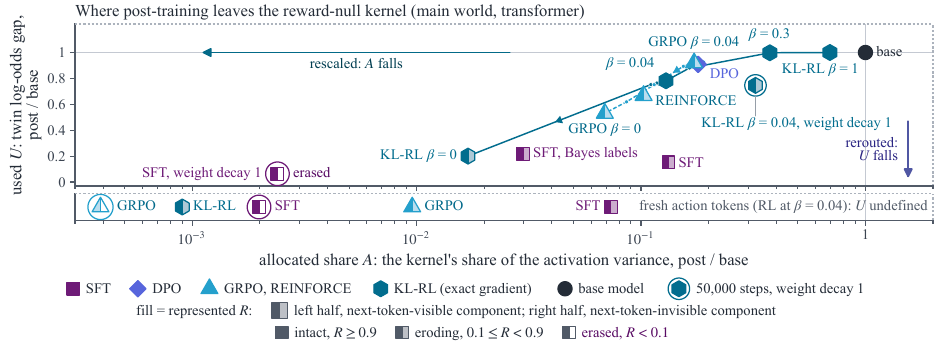}
  \caption{\textbf{Fates as coordinates.} Each marker is one objective in the main world, placed by the share of activation variance it allocates to the kernel ($A$) and by the use its decision makes of the kernel ($U$), both relative to the base model. The two halves of a marker show how much of the next-token-visible and of the next-token-invisible component is still represented ($R$). Lines join the $\beta$ grids of KL-RL (solid) and GRPO (dashed), and DPO coincides with GRPO at $\beta=0.04$. The strip holds the fresh-token objectives, whose decisions never read the kernel. Rescaling moves a marker to the left, rerouting moves it down, and erasure empties it.}
  \label{fig:fates}
\end{figure*}

\paragraph{The filter protects what future decisions need.}
The second constraint comes from how the state is updated. When the observations entangle the two variables, updating the task belief requires the nuisance belief, and the partition that is closed under the filter update is strictly finer than the reward quotient (Appendix~\ref{app:filter}). A recurrent model, whose decision reads a state updated one token at a time, must then carry the reward-null distinctions that future reward-relevant beliefs depend on, whereas a transformer can recompute the belief from the context at the decision.

\paragraph{Free distinctions are left to optimization.}
A distinction that is null for the reward, the reference and the filter is \emph{free}: no policy the objective prefers needs it. Under an anchor in the tied design the free part of the kernel is $\Ninv$. Under supervision, without an anchor, or with a reference that has no preference within a class, it is the whole kernel, less what a recurrent filter needs. What happens to a free distinction is decided by optimization and by pressures outside the objective.
\section{Experiments in Exact Worlds}
\label{sec:results}

\paragraph{Setup.}
Table~\ref{tab:map} lists the worlds, action designs and objectives. The main world is separable, the separable family also contains a point with a slow nuisance chain, and numbers are for the main world unless a world is named. Four-layer transformers and two-layer GRUs are pretrained to the Bayes-optimal loss and post-trained with all parameters for 5{,}000 steps (50{,}000 where stated). All quantities are ratios to the base model at the query position, averaged over three seeds (one for the 50{,}000-step runs). Subspaces, criteria and predictions were fixed before each batch of runs (Appendix~\ref{app:setup}).

\subsection{Compression Is Not Loss}
\label{sec:compress}

Figure~\ref{fig:fates} places each objective in the three coordinates (values in Table~\ref{tab:coords}; other worlds in Appendix~\ref{app:worlds}). Every objective that trains the trunk at $\beta<1$ moves to the left: the kernel's share of the variance at the decision falls, by up to three orders of magnitude, whether or not the information is kept or used. Almost no marker empties. The next-token-visible component stays decodable above 0.9 of the base in every transformer run without weight decay, including the fresh-token objectives whose share falls a thousandfold. What a spectral measurement registers as compression is here a reallocation of variance around information that is still present.

The next-token-invisible component, which nothing in the objective protects, erodes without disappearing: $\Rinv$ ends between 0.57 and 1.00, in an order that follows how far the policy has moved from the reference. The erosion is the trace of drift in shared parameters. It stops when the policy stops moving, as it does under an anchor, and continues slowly without one.

\subsection{The Anchor Decides Use}
\label{sec:anchor}

With an anchor at the coefficients used in practice, trained networks reach the optimum of Proposition~\ref{prop:anchor}. KL-RL, GRPO and DPO reproduce the reference's within-class log-odds (slope $\gamma$ between 0.93 and 1.02 at $\beta\ge0.04$), and in every design the sensitivity of the trained policy to the kernel matches that of $\pistar$ to within 0.01 (Figure~\ref{fig:overview}, II and IV). With a uniform reference the policy never reads the kernel.

Supervised fine-tuning reroutes. After SFT the next-token-visible component is as decodable as before ($\Rvis=0.98$) and the decision has stopped responding to it ($U=0.15$), while the same network's next-token prediction still does. A comparison at matched reward isolates the cause. KL-RL at $\beta=1$ and SFT reach the same reward, the first with $(\Rinv,U,A)=(1.00,1.00,0.69)$ and the second with $(0.75,0.15,0.13)$. The same reward is reached with and without rerouting, and the difference follows from who chooses among equally rewarded answers.

Without the anchor, policy-gradient methods in the separable worlds collapse each class onto one token regardless of the history (Figure~\ref{fig:anchor}, I), while the next-token-visible component stays decodable: rerouting again, now produced by the reward alone. At the policy level the fate switches at a threshold in $\beta$, within $(0.003,0.02]$ for GRPO, whereas $\Rinv$ and $A$ vary smoothly through it. Use is decided by the presence of the anchor, representation and allocation by the distance from the base model. The threshold lies inside the range of current recipes, most of which set $\beta=0$ \citep{yu2025,liu2025,hu2025}.

\subsection{Erasure Needs Pressure Beyond the Objective}
\label{sec:erasure}

None of the above is erasure. To find where erasure begins we apply a stress test, weight decay 1 over 50{,}000 steps, which shrinks whatever no term of the objective holds in place. Under it SFT and the fresh-token objectives take the next-token-invisible component to between 0.001 and 0.012 of the base, at every depth and position (Figure~\ref{fig:overview}, III). Under the same pressure the anchored policy keeps the next-token-visible component intact, still reproduces the reference's log-odds ($\gamma=0.99$), and retains 0.64 of the invisible component. The stress test thus draws the line between kept and erased along whether any term of the objective needs the distinction. The reward's indifference alone erases nothing.

\subsection{A Recurrent State Spreads Freedom and Protection}
\label{sec:recurrent}

\begin{figure}[t]
  \centering
  \includegraphics[width=\columnwidth]{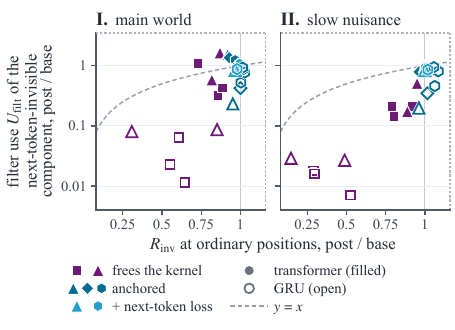}
  \caption{\textbf{Encoded but no longer used.} Use of the next-token-invisible component by the model's own filter against its decodability at ordinary positions, both relative to the base model, for the GRU (open) and the transformer (filled).}
  \label{fig:recurrent}
\end{figure}

A GRU decides from the state it carries along the sequence, so the decision and every other position share one representation. What the objective frees is then lost everywhere, and what it protects is kept everywhere. Under objectives that free the kernel the GRU retains less of the next-token-invisible component than the transformer and loses it at ordinary positions too. Under an anchor in the weight-decay stress test it keeps the component whole (0.99 to 1.02, against 0.64 in the transformer), because the anchored log-odds at the query need the full nuisance belief to arrive through the recurrent state.

Exchanging the kernel component of the state between twins at an ordinary position, and reading the next-token distributions that follow, measures what the model's own filter uses, relative to the exact Bayes filter under the same exchange. The base GRU uses the kernel exactly as Bayes does. After SFT its filter all but stops using the invisible component (0.01 of the base) although half of it is still decodable from the same state: rerouting, now inside the computation (Figure~\ref{fig:recurrent}). An anchor keeps part of that use, and in the entangled world, where the kernel feeds future reward-relevant beliefs, the filter keeps using it under every objective. A recurrent filter keeps what a future reward-relevant belief needs and drops what nothing needs.

\subsection{What Allocation Costs}
\label{sec:allocation}

\begin{figure}[t]
  \centering
  \includegraphics[width=\columnwidth]{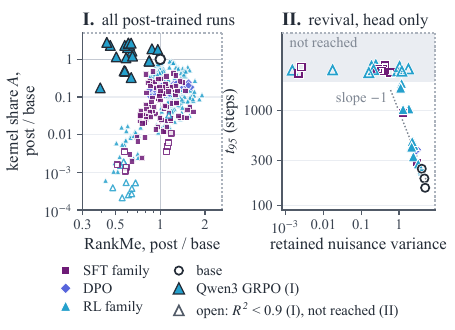}
  \caption{\textbf{Allocation is a third property, and it has a cost.} \textbf{I.} Kernel share against the change in effective rank for all post-trained networks. \textbf{II.} Steps a new linear readout needs to recover the nuisance variable from a frozen network, against the nuisance variance the network retained.}
  \label{fig:allocation}
\end{figure}

Proposition~\ref{prop:gauge} can be applied to a trained network directly. Scaling the kernel's image by 0.1 and compensating at the readout moves the kernel's share twenty- to a hundredfold, with the probe $R^{2}$ and the interchange effect unchanged to numerical precision. Post-training makes the same point: of the 172 runs whose kernel share fell below a tenth of the base, 153 keep $R\ge0.9$ (Figure~\ref{fig:allocation}, I).

Allocation nevertheless has a cost, paid by whoever reads the representation next. We freeze each post-trained network and train a new linear readout on a task that reads the nuisance belief. The steps it needs grow as the nuisance variance the network retained shrinks (rank correlation $-0.93$; II). Once that variance has dropped two-hundredfold, a network that still encodes the nuisance belief linearly ($R^{2}=0.93$) is indistinguishable from an untrained one. Under a fixed adaptation budget, a distinction can remain represented and be practically inaccessible.

\subsection{Optimization Settles What the Reward Leaves Open}
\label{sec:direction}

\begin{figure*}[t]
  \centering
  \includegraphics[width=\textwidth]{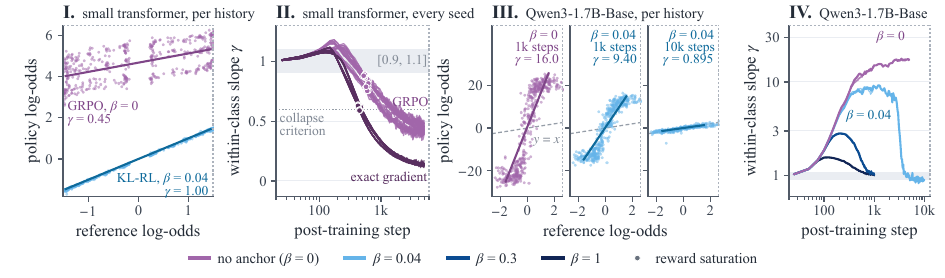}
  \caption{\textbf{With the anchor the reference's log-odds are kept, and without it optimization decides}. \textbf{I.} Within-class log-odds of the policy against the reference, one point per history, in a small transformer: on the diagonal with an anchor, a history-independent band without it. \textbf{II.} Without the anchor the slope $\gamma$ of all 48 runs first rises above 1 and then collapses as the history-independent shift takes over before the reward saturates (dots). \textbf{III, IV.} The same two views in Qwen3-1.7B-Base: without the anchor the log-odds are amplified, with it they return to the base model's.}
  \label{fig:anchor}
\end{figure*}

Without the anchor the objective is silent about tied outputs, and two modes of the policy gradient compete. A history-wise mode amplifies the preference that each history already has within a class, as a policy with separate parameters for every history would. A collective mode shifts the within-class preference of all histories in the same direction, because shared features couple their gradients. At the reference nothing drives this shift, but the reference is an unstable equilibrium for it: any asymmetry of the network grows, and it grows only while the reward is still being learned. Which mode leads when the reward saturates depends on how the representation at the query divides its second moment (Appendix~\ref{app:dynamics}).

In the separable worlds the collective mode wins. The slope $\gamma$ first rises above 1, as the history-wise mode predicts, and the shift overtakes before the reward saturates (Figure~\ref{fig:anchor}, II). Two controls identify the mechanism: training only the head removes the collapse, so the shift is written into the shared representation, and the exact expected gradient collapses as well, so sampling noise does not cause it. The outcome is the same across pretraining and post-training seeds (Appendix~\ref{app:direction}). Where the two modes grow at comparable rates at the base model, in the entangled world and in Mess3, the outcome ranges from sharpening to a partial collapse and is set by the pretrained model. The direction in which a policy leaves the identity is decided while the reward is being learned, by the representation that enters post-training, and later training only reinforces it.
\section{Open Language Models}
\label{sec:llm}

\paragraph{Released checkpoints.}
We repeat the in-context protocol of \citet{balcells2026} on three pairs of base and post-trained checkpoints (Qwen3-1.7B, Qwen3-4B and Qwen3.5-4B; \citealp{yang2025}). The model reads a stream of tokens generated by a hidden Markov model it has never seen, and probes read the belief state and its next-token-invisible component from the residual stream. Post-training leaves both decodable: peak $R^{2}$ changes by at most 0.05 for the belief state and, with three exceptions in the smallest model, by at most 0.06 for the invisible component. Over the same checkpoints the effective rank of the later half of the network falls by 12 to 19\%, while the share of variance carried by the invisible component rises (Appendix~\ref{app:llm}). The spectrum is compressed, decodability does not move, and the compressed directions are not those of the invisible component.

\paragraph{Controlled coarse-class GRPO.}
To intervene on the objective we run the experiment of Section~\ref{sec:anchor} inside a language model. Qwen3-1.7B-Base and Qwen3-4B-Base read an in-context stream from the main world written with six single-token letters, a query marker asks for the next letter, and GRPO with low-rank adapters rewards only the class of the letter. The base model's own next-letter distribution is the reference. The verdict on the anchor repeats (Figure~\ref{fig:anchor}, III and IV). With the anchor the within-class log-odds overshoot and return to the reference's, within 1{,}000 steps at $\beta\ge0.3$ and, in the run we continued, within 10{,}000 at $\beta=0.04$. Without it they are amplified ten to twenty times and no run collapses, in both models and across parameterizations and training lengths (Appendix~\ref{app:llm}). Nothing is erased. The effective rank falls in every run while the kernel's share rises under amplification and falls under the anchor, so allocation again moves independently of information.

The language model thus leaves the identity in the direction opposite to the small transformers. In the terms of Section~\ref{sec:direction}, both systems begin in the history-wise mode. In the separable worlds the collective shift overtakes before the reward saturates, and in the language model it never does: the update amplifies an in-context computation the model already has. What the two systems share is the role of the anchor. With it both return to the reference's log-odds among equally rewarded outputs, and without it neither stays there.
\section{Related Work}
\label{sec:related}

\paragraph{Belief states in sequence models.}
Our object comes from computational mechanics \citep{crutchfield1989,shalizi2001} and from the finding that transformers represent it linearly \citep{shai2024,piotrowski2025}, in context in open language models \citep{balcells2026}. We ask what a second objective, which needs only a quotient of that state, does to what next-token prediction installed.

\paragraph{Post-training and representation change.}
Geometric and spectral studies compare checkpoints before and after post-training \citep{wang2026,li2025,sun2026}, and mechanistic studies find fine-tuning to wrap or enhance existing circuits \citep{jain2024,prakash2024,lee2024}. In all of these the distinctions the objective renders irrelevant are unknown, so a measured change cannot be assigned to loss, rerouting or rescaling. The exact quotient supplies that assignment, and Proposition~\ref{prop:gauge} explains why spectral statistics alone could not.

\paragraph{KL-regularized reinforcement learning and tied outputs.}
That the KL-regularized optimum preserves the reference's proportions among equally rewarded outputs is known \citep{korbak2022,rafailov2023,gxchen2025}. We derive what it implies for the belief state, show that trained networks reach it at practical coefficients, and show what replaces it when the anchor is removed, as in most current recipes \citep{yu2025,liu2025,hu2025}. \citet{shenfeld2026} relate forgetting to the KL from the base policy and argue that policy gradient without a KL term still selects the closest optimal policy. Our unanchored runs leave that policy, in one direction in small transformers and in the other in language models.
\section{Conclusion}
\label{sec:conclusion}

A reward that ignores part of a belief state does not decide what happens to that part. The reward defines which distinctions are decision-equivalent. The anchor and the state update determine which reward-null distinctions remain protected: the anchor keeps what the reference policy distinguishes, and the filter keeps what future reward-relevant beliefs need. Optimization decides the fate of the rest. It usually rescales it, reroutes the decision around it when nothing anchors the decision, and erases it only under sustained pressure and only where no term of the objective needs it. Read through the three coordinates, a compressed spectrum stops being evidence of loss, and the question of whether post-training compresses a representation gives way to the question of which causal quotient of the belief state it implements.
\section*{Limitations}
\label{sec:limitations}

The exact ground truth that makes the three fates identifiable exists only in small hidden Markov worlds, with a one-step decision whose action is not fed back into the context. Whether a multi-step task, in which the policy also changes which histories are visited, moves the quotient or only the support of the belief state remains to be studied. The language-model experiments cover three pairs of released checkpoints from one model family and one controlled objective on in-context synthetic streams, with models of 1.7 and 4 billion parameters. They show that the dissociation and the role of the anchor carry over to pretrained language models, and the size of either effect in post-training on natural tasks is unmeasured. Represented information is measured with linear probes, so a distinction that survives only in a nonlinear code is counted as lost. Small transformers and language models leave the anchor identity in opposite directions, and neither parameterization nor training budget accounts for the difference. What makes amplification the cheaper direction in a pretrained language model is open.

\bibliography{references}

\begin{thebibliography}{24}
\providecommand{\natexlab}[1]{#1}

\bibitem[{Balcells et~al.(2026)Balcells, Lee, Rastogi, Riechers, Shai, and
  Poncini}]{balcells2026}
Daniel Balcells, Andrew~Jun Lee, Chirag Rastogi, Paul~M. Riechers, Adam Shai,
  and Xavier Poncini. 2026.
\newblock \href {https://arxiv.org/abs/2609.17376} {Large language models
  develop belief state geometry in-context}.
\newblock \emph{Preprint}, arXiv:2609.17376.

\bibitem[{Crutchfield and Young(1989)}]{crutchfield1989}
James~P. Crutchfield and Karl Young. 1989.
\newblock \href {https://doi.org/10.1103/PhysRevLett.63.105} {Inferring
  statistical complexity}.
\newblock \emph{Physical Review Letters}, 63(2):105--108.

\bibitem[{Garrido et~al.(2023)Garrido, Balestriero, Najman, and
  LeCun}]{garrido2023}
Quentin Garrido, Randall Balestriero, Laurent Najman, and Yann LeCun. 2023.
\newblock \href {https://proceedings.mlr.press/v202/garrido23a.html} {{RankMe}:
  Assessing the downstream performance of pretrained self-supervised
  representations by their rank}.
\newblock In \emph{Proceedings of the 40th International Conference on Machine
  Learning}, volume 202 of \emph{Proceedings of Machine Learning Research},
  pages 10929--10974. PMLR.

\bibitem[{Geiger et~al.(2021)Geiger, Lu, Icard, and Potts}]{geiger2021}
Atticus Geiger, Hanson Lu, Thomas Icard, and Christopher Potts. 2021.
\newblock \href
  {https://proceedings.neurips.cc/paper_files/paper/2021/hash/4f5c422f4d49a5a807eda27434231040-Abstract.html}
  {Causal abstractions of neural networks}.
\newblock In \emph{Advances in Neural Information Processing Systems},
  volume~34, pages 9574--9586. Curran Associates, Inc.

\bibitem[{Guo et~al.(2025)Guo, Yang, Zhang, Song, Wang et~al.}]{guo2025}
Daya Guo, Dejian Yang, Haowei Zhang, Junxiao Song, Peiyi Wang, et~al. 2025.
\newblock \href {https://doi.org/10.1038/s41586-025-09422-z} {{DeepSeek-R1}
  incentivizes reasoning in {LLMs} through reinforcement learning}.
\newblock \emph{Nature}, 645(8081):633--638.

\bibitem[{GX-Chen et~al.(2026)GX-Chen, Prakash, Guo, Fergus, and
  Ranganath}]{gxchen2025}
Anthony GX-Chen, Jatin Prakash, Jeff Guo, Rob Fergus, and Rajesh Ranganath.
  2026.
\newblock \href
  {https://proceedings.iclr.cc/paper_files/paper/2026/hash/5306f6ce1082d29e4968acdacab84fd7-Abstract-Conference.html}
  {{KL}-regularized reinforcement learning for generative modelling is designed
  to mode collapse}.
\newblock In \emph{The Fourteenth International Conference on Learning
  Representations}, pages 50665--50693.

\bibitem[{Hu et~al.(2025)Hu, Zhang, Han, Jiang, Zhang, and Shum}]{hu2025}
Jingcheng Hu, Yinmin Zhang, Qi~Han, Daxin Jiang, Xiangyu Zhang, and Heung-Yeung
  Shum. 2025.
\newblock \href {https://doi.org/10.52202/085713-5418} {{Open-Reasoner-Zero}:
  An open source approach to scaling up reinforcement learning on the base
  model}.
\newblock In \emph{Advances in Neural Information Processing Systems},
  volume~38, pages 162239--162262. Curran Associates, Inc.

\bibitem[{Jain et~al.(2024)Jain, Kirk, Lubana, Dick, Tanaka, Rockt{\"a}schel,
  Grefenstette, and Krueger}]{jain2024}
Samyak Jain, Robert Kirk, Ekdeep~Singh Lubana, Robert~P. Dick, Hidenori Tanaka,
  Tim Rockt{\"a}schel, Edward Grefenstette, and David Krueger. 2024.
\newblock \href {https://openreview.net/forum?id=A0HKeKl4Nl} {Mechanistically
  analyzing the effects of fine-tuning on procedurally defined tasks}.
\newblock In \emph{The Twelfth International Conference on Learning
  Representations}.

\bibitem[{Korbak et~al.(2022)Korbak, Perez, and Buckley}]{korbak2022}
Tomasz Korbak, Ethan Perez, and Christopher Buckley. 2022.
\newblock \href {https://doi.org/10.18653/v1/2022.findings-emnlp.77} {{RL} with
  {KL} penalties is better viewed as {B}ayesian inference}.
\newblock In \emph{Findings of the Association for Computational Linguistics:
  EMNLP 2022}, pages 1083--1091, Abu Dhabi, United Arab Emirates. Association
  for Computational Linguistics.

\bibitem[{Lee et~al.(2024)Lee, Bai, Pres, Wattenberg, Kummerfeld, and
  Mihalcea}]{lee2024}
Andrew Lee, Xiaoyan Bai, Itamar Pres, Martin Wattenberg, Jonathan~K.
  Kummerfeld, and Rada Mihalcea. 2024.
\newblock \href {https://proceedings.mlr.press/v235/lee24a.html} {A mechanistic
  understanding of alignment algorithms: A case study on {DPO} and toxicity}.
\newblock In \emph{Proceedings of the 41st International Conference on Machine
  Learning}, volume 235 of \emph{Proceedings of Machine Learning Research},
  pages 26361--26378. PMLR.

\bibitem[{Li et~al.(2025)Li, Agrawal, Ghosh, Teru, Santoro, Lajoie, and
  Richards}]{li2025}
Melody Li, Kumar~Krishna Agrawal, Arna Ghosh, Komal Teru, Adam Santoro,
  Guillaume Lajoie, and Blake Richards. 2025.
\newblock \href {https://doi.org/10.52202/085713-1829} {Tracing the
  representation geometry of language models from pretraining to
  post-training}.
\newblock In \emph{Advances in Neural Information Processing Systems},
  volume~38, pages 54691--54724. Curran Associates, Inc.

\bibitem[{Liu et~al.(2025)Liu, Chen, Li, Qi, Pang, Du, Lee, and Lin}]{liu2025}
Zichen Liu, Changyu Chen, Wenjun Li, Penghui Qi, Tianyu Pang, Chao Du, Wee~Sun
  Lee, and Min Lin. 2025.
\newblock \href {https://openreview.net/forum?id=5PAF7PAY2Y} {Understanding
  {R1-Zero}-like training: A critical perspective}.
\newblock In \emph{Second Conference on Language Modeling}.

\bibitem[{Piotrowski et~al.(2025)Piotrowski, Riechers, Filan, and
  Shai}]{piotrowski2025}
Mateusz Piotrowski, Paul~M. Riechers, Daniel Filan, and Adam Shai. 2025.
\newblock \href {https://proceedings.mlr.press/v267/piotrowski25a.html}
  {Constrained belief updates explain geometric structures in transformer
  representations}.
\newblock In \emph{Proceedings of the 42nd International Conference on Machine
  Learning}, volume 267 of \emph{Proceedings of Machine Learning Research},
  pages 49399--49419. PMLR.

\bibitem[{Prakash et~al.(2024)Prakash, Rott~Shaham, Haklay, Belinkov, and
  Bau}]{prakash2024}
Nikhil Prakash, Tamar Rott~Shaham, Tal Haklay, Yonatan Belinkov, and David Bau.
  2024.
\newblock \href {https://openreview.net/forum?id=8sKcAWOf2D} {Fine-tuning
  enhances existing mechanisms: A case study on entity tracking}.
\newblock In \emph{The Twelfth International Conference on Learning
  Representations}.

\bibitem[{Rafailov et~al.(2023)Rafailov, Sharma, Mitchell, Manning, Ermon, and
  Finn}]{rafailov2023}
Rafael Rafailov, Archit Sharma, Eric Mitchell, Christopher~D. Manning, Stefano
  Ermon, and Chelsea Finn. 2023.
\newblock \href {https://doi.org/10.52202/075280-2338} {Direct preference
  optimization: Your language model is secretly a reward model}.
\newblock In \emph{Advances in Neural Information Processing Systems},
  volume~36, pages 53728--53741. Curran Associates, Inc.

\bibitem[{Shai et~al.(2024)Shai, Marzen, Teixeira, Gietelink~Oldenziel, and
  Riechers}]{shai2024}
Adam~S. Shai, Sarah~E. Marzen, Lucas Teixeira, Alexander Gietelink~Oldenziel,
  and Paul~M. Riechers. 2024.
\newblock \href {https://doi.org/10.52202/079017-2387} {Transformers represent
  belief state geometry in their residual stream}.
\newblock In \emph{Advances in Neural Information Processing Systems},
  volume~37, pages 75012--75034. Curran Associates, Inc.

\bibitem[{Shalizi and Crutchfield(2001)}]{shalizi2001}
Cosma~Rohilla Shalizi and James~P. Crutchfield. 2001.
\newblock \href {https://doi.org/10.1023/A:1010388907793} {Computational
  mechanics: Pattern and prediction, structure and simplicity}.
\newblock \emph{Journal of Statistical Physics}, 104(3--4):817--879.

\bibitem[{Shao et~al.(2024)Shao, Wang, Zhu, Xu, Song, Bi, Zhang, Zhang, Li, Wu,
  and Guo}]{shao2024}
Zhihong Shao, Peiyi Wang, Qihao Zhu, Runxin Xu, Junxiao Song, Xiao Bi, Haowei
  Zhang, Mingchuan Zhang, Y.~K. Li, Y.~Wu, and Daya Guo. 2024.
\newblock \href {https://arxiv.org/abs/2402.03300} {{DeepSeekMath}: Pushing the
  limits of mathematical reasoning in open language models}.
\newblock \emph{Preprint}, arXiv:2402.03300.

\bibitem[{Shenfeld et~al.(2026)Shenfeld, Pari, and Agrawal}]{shenfeld2026}
Idan Shenfeld, Jyothish Pari, and Pulkit Agrawal. 2026.
\newblock \href {https://openreview.net/forum?id=7HNRYT4V44} {{RL}'s razor: Why
  online reinforcement learning forgets less}.
\newblock In \emph{The Fourteenth International Conference on Learning
  Representations}.

\bibitem[{Sun et~al.(2026)Sun, Dong, Qiao, Lin, Zhang, and Rajmohan}]{sun2026}
Lihao Sun, Hang Dong, Bo~Qiao, Qingwei Lin, Dongmei Zhang, and Saravan
  Rajmohan. 2026.
\newblock \href {https://doi.org/10.18653/v1/2026.acl-long.1237} {{LLM}
  reasoning as trajectories: Step-specific representation geometry and
  correctness signals}.
\newblock In \emph{Proceedings of the 64th Annual Meeting of the {A}ssociation
  for {C}omputational {L}inguistics (Volume 1: Long Papers)}, pages
  26872--26887, San Diego, California, United States. Association for
  Computational Linguistics.

\bibitem[{Wang et~al.(2026)Wang, Yang, Wang, and Zhang}]{wang2026}
Juntong Wang, Shengkun Yang, Xiyuan Wang, and Muhan Zhang. 2026.
\newblock \href {https://arxiv.org/abs/2608.01835} {Rewriting or reweighting?
  {A} geometric account in language models}.
\newblock \emph{Preprint}, arXiv:2608.01835.

\bibitem[{Xu et~al.(2020)Xu, Zhao, Song, Stewart, and Ermon}]{xu2020}
Yilun Xu, Shengjia Zhao, Jiaming Song, Russell Stewart, and Stefano Ermon.
  2020.
\newblock \href {https://openreview.net/forum?id=r1eBeyHFDH} {A theory of
  usable information under computational constraints}.
\newblock In \emph{International Conference on Learning Representations}.

\bibitem[{Yang et~al.(2025)Yang, Li, Yang, Zhang, Hui, Zheng, Yu, Gao, Huang,
  Lv et~al.}]{yang2025}
An~Yang, Anfeng Li, Baosong Yang, Beichen Zhang, Binyuan Hui, Bo~Zheng, Bowen
  Yu, Chang Gao, Chengen Huang, Chenxu Lv, et~al. 2025.
\newblock \href {https://arxiv.org/abs/2505.09388} {{Qwen3} technical report}.
\newblock \emph{Preprint}, arXiv:2505.09388.

\bibitem[{Yu et~al.(2025)Yu, Zhang, Zhu, Yuan, Zuo, Yue, Dai, Fan, Liu, Liu,
  Liu, Liu, Lin, Lin, Ma, Sheng, Tong, Zhang, Zhang, Zhang, Zhang, Zhu, Zhu,
  Chen, Chen, Wang, Yu, Song, Wei, Zhou, Liu, Ma, Zhang, Yan, Wu, and
  Wang}]{yu2025}
Qiying Yu, Zheng Zhang, Ruofei Zhu, Yufeng Yuan, Xiaochen Zuo, Yu~Yue, Weinan
  Dai, Tiantian Fan, Gaohong Liu, Juncai Liu, Lingjun Liu, Xin Liu, Haibin Lin,
  Zhiqi Lin, Bole Ma, Guangming Sheng, Yuxuan Tong, Chi Zhang, Mofan Zhang, and
  17 others. 2025.
\newblock \href {https://doi.org/10.52202/085713-3775} {{DAPO}: An open-source
  {LLM} reinforcement learning system at scale}.
\newblock In \emph{Advances in Neural Information Processing Systems},
  volume~38, pages 113222--113244. Curran Associates, Inc.

\end{thebibliography}

\appendix
\setcounter{topnumber}{3}\setcounter{dbltopnumber}{3}\setcounter{totalnumber}{5}
\renewcommand{\topfraction}{0.85}\renewcommand{\dbltopfraction}{0.8}
\renewcommand{\textfraction}{0.12}
\renewcommand{\floatpagefraction}{0.7}\renewcommand{\dblfloatpagefraction}{0.7}
\section{Theory}
\label{app:theory}

\subsection{The anchor identity and what it protects}

\paragraph{Proof of Proposition~\ref{prop:anchor}.}
For a fixed history the objective is concave in $\pi(\cdot\mid h)$ on the simplex, and its unique maximizer is $\pistar(a\mid h)=\piref(a\mid h)\,e^{q(h)_{a}/\beta}/Z_{\beta}(h)$. Taking logarithms, $\log\pistar(a\mid h)-\log\pistar(a'\mid h)=\log\piref(a\mid h)-\log\piref(a'\mid h)+(q(h)_{a}-q(h)_{a'})/\beta$. The last term vanishes when the two actions are tied, which is (i). When $b(h)\simR b(h')$ the last term is the same for the two histories and cancels in their difference, which is (ii). \hfill$\square$

Statement (i) extends to any objective in which the signal for an action enters only through its reward, such as a temperature that depends on the history, an advantage normalized within a group, or a shaping that depends only on the class. The group-wise standard-deviation normalization of GRPO therefore leaves (i) intact.

\paragraph{Protected and free directions.}
In the tied design the action set is the token alphabet, $q(b)_{a}=\bk_{\cls(a)}$, and $\piref(\cdot\mid h)=b(h)^{\top}E$. The optimum is a function of $(\bk,b^{\top}E)$, and by (i) its conditional distribution within a class equals that of $b^{\top}E$ for every $\beta$. A tangent direction $d$ leaves $\pistar$ unchanged for every $\beta\in(0,\infty]$ exactly when it changes neither $\bk$ nor $b^{\top}E$, that is, when $d\in\Ninv$. In the main world $\Ninv$ has three dimensions and carries 11 to 12\% of the variance of the belief cloud, and $\Nvis$ has the other three dimensions of the kernel. When each class keeps a single action token there are no ties, only (ii) remains, and the sensitivity of $\pistar$ to the kernel in total variation follows a curve in $\beta$ that we use as a quantitative test (Figure~\ref{fig:overview}, IV). When the action tokens are new, the reference is uniform and $\pistar$ ignores the whole kernel.

\paragraph{Rescaling at the output.}
Write $\ell=C\log\pi$ for centered log-probabilities. At the optimum $\ell^{*}_{\beta}=\ell_{\mathrm{ref}}+Cq/\beta$. The variance of $\ell^{*}_{\beta}$ within a fiber equals that of $\ell_{\mathrm{ref}}$ for every $\beta$, and the total variance is $a+2c/\beta+c'/\beta^{2}$ with constants set by the world. The kernel's share of the output variance in the main world falls from 0.30 at $\beta=\infty$ to $3\times10^{-3}$ at $\beta=0.04$ and $2\times10^{-6}$ at $\beta=0.001$, with its log-odds unchanged. Trained networks follow this curve for $\beta\ge0.1$ and stop near 0.01 for smaller $\beta$: they amplify the class logits until the reward saturates and no further (Figure~\ref{fig:app-spectral}).

\paragraph{The supervised optimum.}
If labels are drawn as $a\sim p_{\mathrm{demo}}(\cdot\mid c)$ given the true class $c$, the minimizer of the expected cross-entropy at belief $b$ is $\sum_{k}\bk_{k}\,p_{\mathrm{demo}}(a\mid k)$. It depends on $b$ only through $\bk$ and is independent of $\piref$. The same holds for labels given by the Bayes action. At matched reward the supervised optimum lies 19 to 33 times farther from the reference in KL than $\pistar$ when the demonstrator names one canonical token per class, and the trained networks give 22 to 24.

\subsection{Filter closure}
\label{app:filter}

Write $b_{w}$ for the belief $b$ after the filter has been updated on a token sequence $w$. Let $b\simF b'$ when $q(b_{w})=q(b'_{w})$ for every $w$, the empty sequence included: the reward vectors of the two beliefs agree now and after every continuation. $\simF$ is the coarsest partition that refines $\simR$ and is closed under the filter update. Take binary $c$ and $f$ with observation $o=c\oplus f$, and two beliefs with $c$ uniform and $f$ known to be 0 or 1. They are reward-equivalent. After observing $o=0$ the first implies $c=0$ and the second $c=1$. A state that keeps only $\bk$ cannot be updated correctly, so $\simF$ is strictly finer than $\simR$. In the separable worlds the class belief updates on its own and $\simF=\simR$ on reachable beliefs. In the entangled world one step of the exact filter after exchanging the kernel component between twins changes the reward-relevant statistic by 0.14 on average and flips the best class in 17\% of cases.

\subsection{The spectrum}

\paragraph{Proof of Proposition~\ref{prop:gauge}.}
The composed map is unchanged because $g\circ G^{-1}\circ G=g$. A probe $f$ on $x$ and the probe $f\circ G^{-1}$ on $Gx$ make identical predictions, so a probe family closed under $G$ attains the same optimum on both. Unregularized least squares is such a family. Ridge regression is not. An interchange that replaces the component of $x$ in a subspace $S$ along a complement $S'$ by that of a donor corresponds to the interchange on $GS$ along $GS'$, and yields the same output. For the covariance, write $\Sigma=LL^{\top}$ and a target $\Sigma'=L'L'^{\top}$ with factors of equal rank and take $G$ with $GL=L'$. \hfill$\square$

A linear probe with a penalty on its weights reads a change of allocation as a loss of information: when the kernel component of an exactly embedded belief cloud is scaled by 0.03 and the readout compensated, the least-squares $R^{2}$ of the kernel stays at 0.954 and the ridge $R^{2}$ falls to 0.139. This is why all probes here whiten their inputs and use no penalty. The converse of the proposition is not a theorem either: projecting a direction out of the activations raises or lowers the effective rank depending on that direction's share of the spectrum.

\subsection{Tied actions under shared parameters}
\label{app:dynamics}

Index the two tied tokens of class $k$ by $j\in\{0,1\}$, let $d_{k}(h)$ be their log-odds, $p_{k}(h)$ the probability of the class, and $\tilde A_{k}(h)$ the class-level advantage. The gradient of the expected reward with respect to the logits gives the within-class source $s_{k}=p_{k}\tilde A_{k}\tanh(d_{k}/2)$.

\paragraph{Separate parameters.}
With one logit vector per history, gradient flow without an anchor gives $\dot d_{k}=s_{k}$. In the class the policy favors, $\tilde A_{k}>0$ and $|d_{k}|$ grows with its sign preserved: the token the reference prefers in this history is amplified, and the slope $\gamma$ against the reference grows about logarithmically in time. Natural policy gradient moves along the tilt family and keeps $\gamma=1$. With an anchor the slope overshoots and returns to 1.

\paragraph{Shared parameters.}
With features $x(h)$ and a linear head, $\dot d_{k}(h)=\eta\,\Exp_{h'}[K(h,h')\,s_{k}(h')]$ with $K(h,h')=1+x(h)\cdot x(h')$. In a separable world the class belief and the nuisance belief are independent, the classes are exchangeable and the two nuisance symbols can be swapped. At the reference $p_{k}\tilde A_{k}$ depends only on the class belief and $\tanh(d_{k}/2)$ only on the nuisance belief, and is odd under the swap. Three consequences follow. The source has zero projection on every feature that depends on the class belief alone, constants included, and on every feature that depends on the nuisance belief alone, so only interaction features are driven. Linearizing around the reference, $d_{k}=d_{\mathrm{ref}}+v_{k}\cdot x$, gives $\dot v_{k}=\eta\,\Exp[w_{k}xx^{\top}]v_{k}$ with $w_{k}=p_{k}\tilde A_{k}\cdot\tfrac12\operatorname{sech}^{2}(d_{k}/2)$. This operator vanishes on nuisance features and has positive eigenvalues on class features. A shift of $d_{k}$ that is gated by the class belief and constant across the nuisance belief is therefore undriven and linearly unstable: the reference is a symmetric unstable equilibrium of this collective mode. Finally, the classes couple through the class probabilities with a positive coefficient, so the combination in which all classes choose the same symbol grows fastest.

The collective mode grows from a seed $\epsilon$ as $\epsilon\,e^{\Lambda(t)}$, where $\Lambda$ accumulates only while the class-level advantage is unsaturated, and the history-wise mode grows linearly at a rate set by the variance of interaction features. Which one dominates when the reward saturates depends on how the representation at the query divides its second moment. In the small transformers the collective rate exceeds the history-wise rate by a factor of $10^{4}$ at the base model. The seed is the network's own asymmetry, and sampling noise decides only where that asymmetry is small (Appendix~\ref{app:direction}).

\paragraph{The anchor.}
The KL term acts on a collective shift $e$ with force $-\beta e\,\Exp[p_{k}\,\sigma(d_{\mathrm{ref}}+e)\,\sigma(-d_{\mathrm{ref}}-e)]$. Histories in which the reference strongly disfavors the winning token contribute weights of order one, so the restoring force is of order $\beta e$. With separate parameters the corresponding force on a collapsed history is exponentially small. A small anchor therefore undoes a collapse in a shared-parameter model, with the intercept recovering before the slope.

\section{Worlds, Models and Training}
\label{app:setup}

\paragraph{Worlds.}
Transitions stay in the current state with probability $1-x$ and move uniformly otherwise. In the separable family the task chain has $x=0.1$ and emits its own symbol with probability 0.85. The nuisance chain has three states and two symbols: two states emit their own symbol with probability $\alpha_{F}$ and the third emits either with probability one half, so the emission has rank two and one direction of the nuisance belief is invisible to the next token. The main world uses $x_{F}=0.15$ and $\alpha_{F}=0.95$, a second point $x_{F}=0.05$ and $\alpha_{F}=0.99$, and the slow world $x_{F}=0.02$ and $\alpha_{F}=0.85$. The control replaces the nuisance emission by an invertible one, which makes the whole kernel visible. The entangled world emits a pair: the sum of the two variables modulo three with fidelity 0.85, and a copy of the task variable with fidelity 0.75, which is needed to make the task variable identifiable. The Mess3 world uses the parameters of \citet{shai2024} with the classes $\{A\}$ and $\{B,C\}$, which leaves a one-dimensional kernel. A query token is inserted after an observation with probability 0.05 from the eighth observation on, and the next observation follows it, so the hidden state does not depend on the query or on the action.

\paragraph{Models and pretraining.}
Transformers have four pre-norm blocks of width 128 with four heads, RMS normalization and rotary positions. GRUs have two layers of width 128. Both are trained on fresh streams of 64 tokens for 40{,}000 steps with AdamW. All base models end within $5\times10^{-4}$ nat of the Bayes-optimal loss, and the belief state and both kernel components are linearly decodable above the baselines given by an untrained network and by the last eight tokens.

\paragraph{Post-training.}
The standard configuration is 5{,}000 steps of AdamW at learning rate $3\times10^{-5}$ on batches of 256 streams with all parameters trainable and the action head tied to the unembedding. SFT uses the label of a demonstrator who knows the class and names its canonical token, or the Bayes action. DPO uses preference pairs from a Bradley--Terry model on the reward, for which the optimum is the same tilt as for KL-regularized reinforcement learning. GRPO uses groups of eight sampled actions, the standard-deviation normalization of the advantage, and the exact KL to the reference. Dr.\ GRPO \citep{liu2025} is GRPO without the division by the standard deviation. A supervised control (SFT on $\pistar$) minimizes the cross-entropy to the exact anchored optimum. KL-RL uses the exact expected reward and its gradient. Auxiliary conditions add a next-token loss at ordinary positions, weight decay, longer training, low-rank adapters, an untied head, or a frozen trunk.

\paragraph{Pre-registration.}
The kernel subspaces, the criteria for each fate and the predictions for each batch of runs were committed to the repository before that batch was launched. Two directional predictions were overturned by the data and are reported as findings: the prediction from separate-parameter dynamics that unanchored policy gradient sharpens tied outputs (Section~\ref{sec:direction}), and the prediction from the optimum that a recurrent model retains more of the free kernel than a transformer (Section~\ref{sec:recurrent}).

\section{Measurements}
\label{app:measure}

\paragraph{Scope.}
$R$ and $A$ are computed for the kernel and for each of its two components. $U$ at the decision is computed from twins, which differ along the whole kernel, and is a property of the kernel. The use of a single component is measured by exchanging that component's image in the activations, at the decision or in the filter-use test below.

\paragraph{Represented.}
Targets are tangent coordinates of the exact belief projected on the kernel, on its two components and on the reward-relevant subspace. The probe is least squares on inputs whitened within each training fold, with five folds split by sequence and a pooled held-out $R^{2}$. Every evaluation checks that $R^{2}$ is unchanged under a random invertible map of the activations. We also report a probe frozen at the base model, whose failure on a post-trained model with an intact refit probe indicates re-encoding.

\paragraph{Used.}
The twin set contains 16{,}384 pairs of streams with queries at the same positions. In the separable worlds twins share the task chain and resample the nuisance chain. In the entangled world the second stream rotates the sum channel, and in Mess3 it mirrors $B$ and $C$. In every case the class marginal agrees between twins to $10^{-9}$. $U$ is the mean norm of the difference of centered log-policies at the first query of each stream (4{,}096 pairs), and its counterpart for next-token prediction reads the language-model head. For the tied design the slope $\gamma$ and intercept come from a least-squares fit of the policy's within-class log-odds on the reference's over histories, restricted to the class with the largest policy mass. The interchange intervention replaces the component of the residual stream in a subspace by the twin's and reports the fraction of the twin effect it transfers, against random subspaces of the same dimension. The subspace is the image of the kernel under a linear fit of the activations on the belief, fitted at the base model or refit at the model under test.

\paragraph{Allocated.}
The share is the variance explained by the kernel coordinates in a regression of the activations on all belief coordinates, divided by the total variance. The effective rank is RankMe on the same activations.

\paragraph{Filter use.}
At an ordinary position $t'$ the kernel component of the state is replaced by that of a twin. For the GRU the whole recurrent state is patched and the recurrence rerun. For the transformer the residual stream is patched at all positions up to $t'$ and all depths, and later positions attend to it. $\Ufilt(k)$ is the KL between the patched and the original next-token distribution $k$ positions later, summed over pairs, divided by the same quantity for the exact Bayes filter. Patching a single layer or position is undone by the layers and positions left untouched, which is why the whole state is used.

\section{Additional Results in Exact Worlds}
\label{app:worlds}

\begin{figure*}[p]
  \centering
  \includegraphics[width=\textwidth]{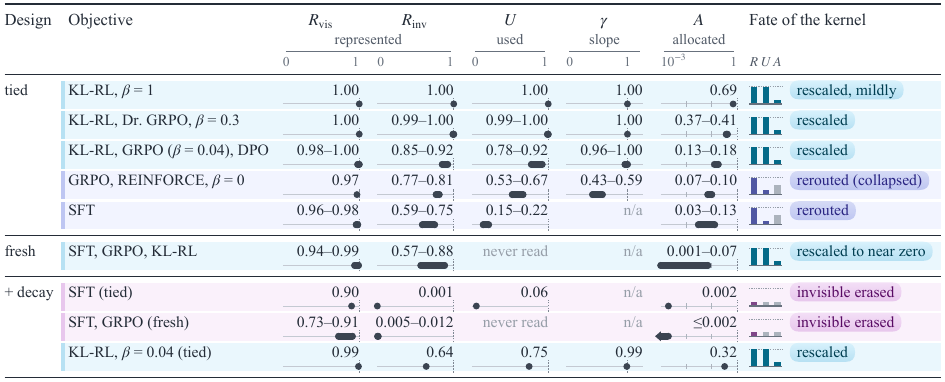}
  \captionof{table}{\textbf{Where each objective leaves the reward-null kernel} (main world, transformer; means over three base seeds or ranges over the objectives of a row, one seed in the last block). Each cell prints its value and marks it on a track. $\Rvis$, $\Rinv$: decodability of the next-token-visible and of the next-token-invisible component. $U$: effect of the kernel on the decision. $\gamma$: slope of the policy's within-class log-odds against the reference's, with a mark at 1. $A$: variance share of the kernel, on a logarithmic track. All are ratios to the base model (dashed) after 5{,}000 steps, and the last block adds weight decay 1 over 50{,}000 steps. The bars beside each fate are its signature in $R$, $U$ and $A$, as in Figure~\ref{fig:overview}.}
  \label{tab:coords}
  \par\vspace{1.6em}
  \includegraphics[width=\textwidth]{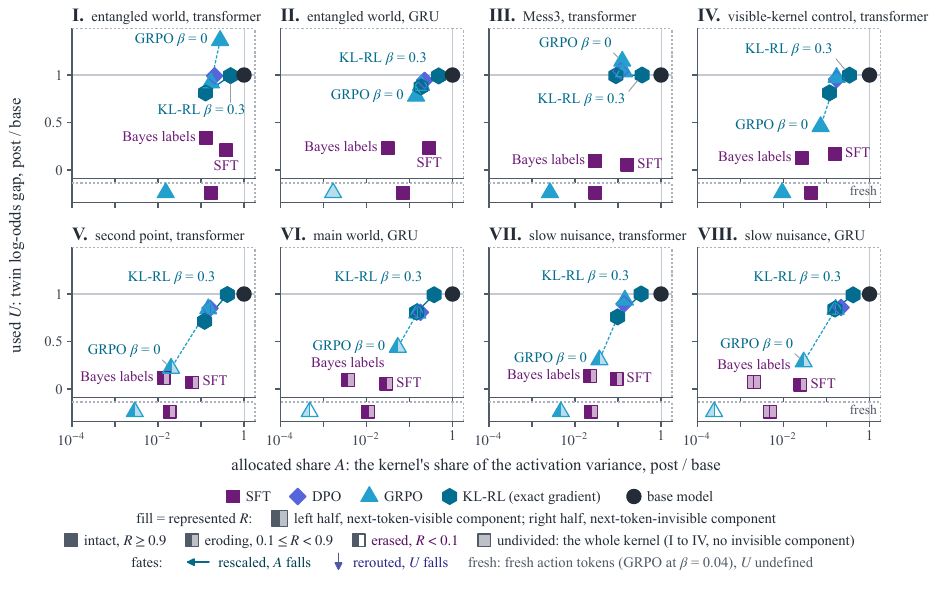}
  \captionof{figure}{\textbf{The coordinates of Figure~\ref{fig:fates} in the other worlds and in the GRU.} Allocated share against use at the decision position, with the fill of a marker showing what is still represented, for eight combinations of world and architecture. Dashed lines join GRPO at $\beta=0.04$ and $0$, solid lines KL-RL at $\beta=0.3$ and $0.04$.}
  \label{fig:app-fates}
\end{figure*}

\begin{figure*}[p]
  \centering
  \includegraphics[width=\textwidth]{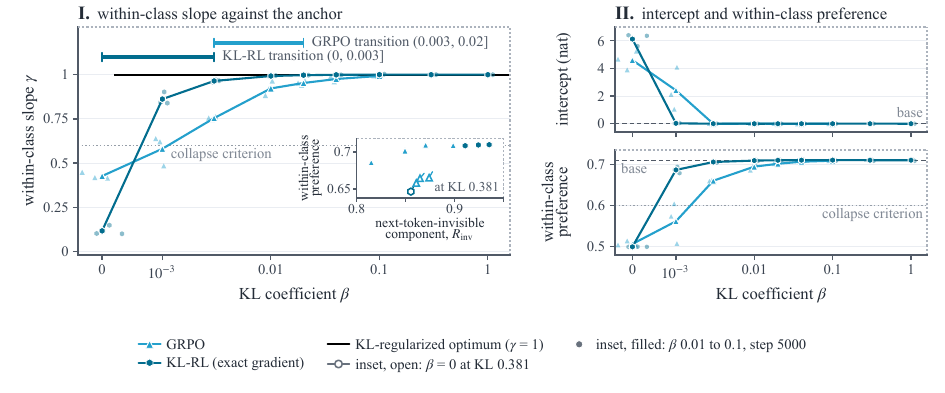}
  \caption{\textbf{The anchor threshold.} Slope, intercept and within-class preference over $\beta$ for GRPO and KL-RL, with the transition intervals. Inset: at matched KL from the reference, unanchored runs already differ from anchored ones in their within-class preference and not in the decodability of the invisible component.}
  \label{fig:app-anchor}
  \par\vspace{1.6em}
  \begin{minipage}[t]{\columnwidth}
    \centering
    \includegraphics[width=\linewidth]{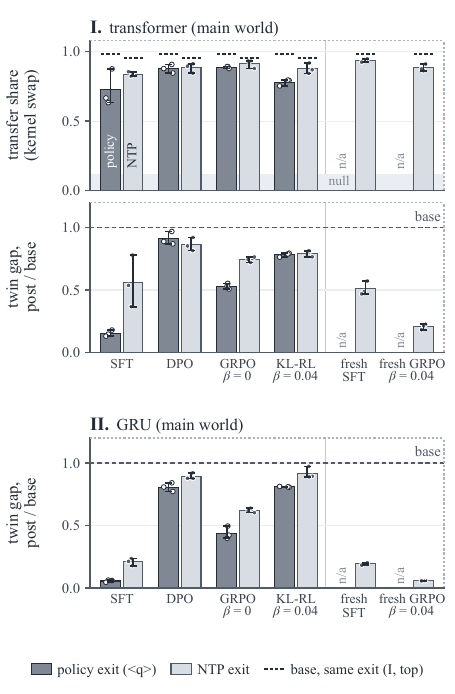}
    \caption{\textbf{Two exits of one network.} Transfer of the twin effect through the kernel subspace and twin effect relative to the base, at the policy and at the next-token head.}
    \label{fig:app-exits}
  \end{minipage}\hfill
  \begin{minipage}[t]{\columnwidth}
    \centering
    \includegraphics[width=\linewidth]{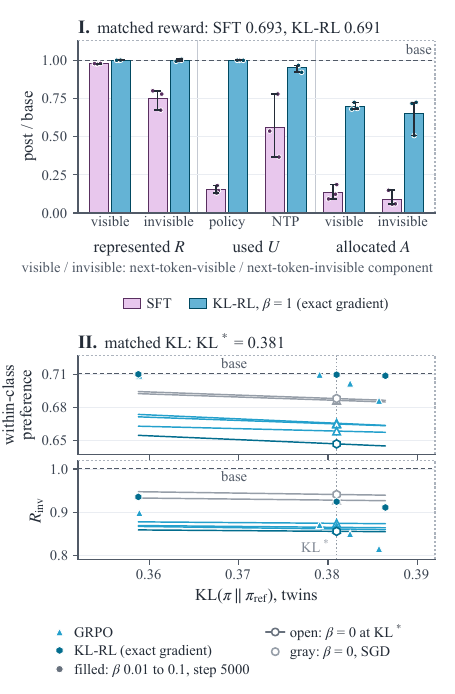}
    \caption{\textbf{Matched reward and matched KL.} Coordinates of SFT and KL-RL at the same reward, and of unanchored and anchored reinforcement learning at the same KL from the reference.}
    \label{fig:app-matched}
  \end{minipage}
\end{figure*}

\begin{figure*}[t]
  \centering
  \includegraphics[width=\textwidth]{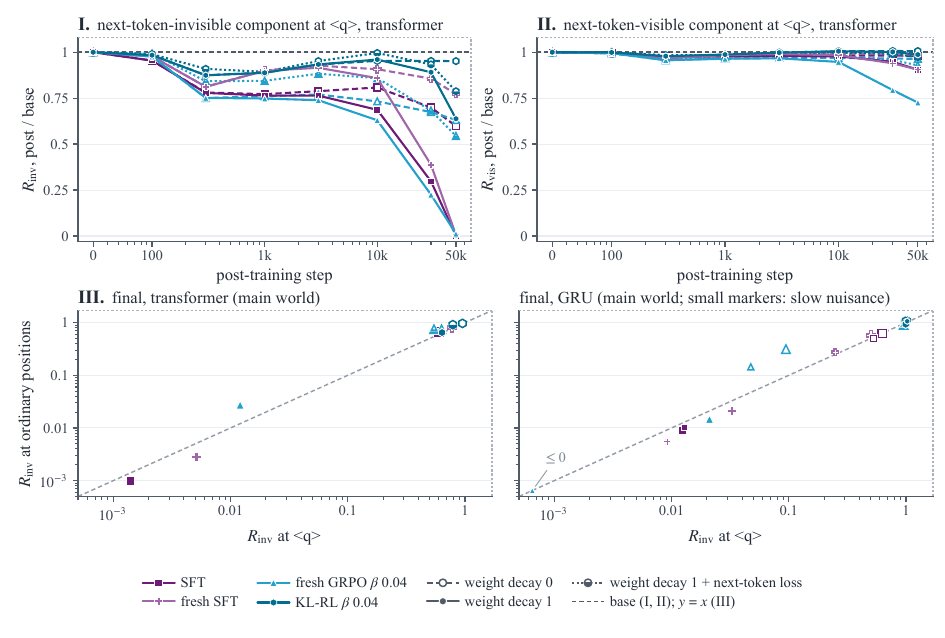}
  \caption{\textbf{Erasure in the weight-decay corner.} Trajectories of both kernel components in the transformer, and ordinary positions against the query position at the end of training for the transformer and the GRU.}
  \label{fig:app-erasure}
\end{figure*}

\begin{figure*}[p]
  \centering
  \includegraphics[width=\textwidth]{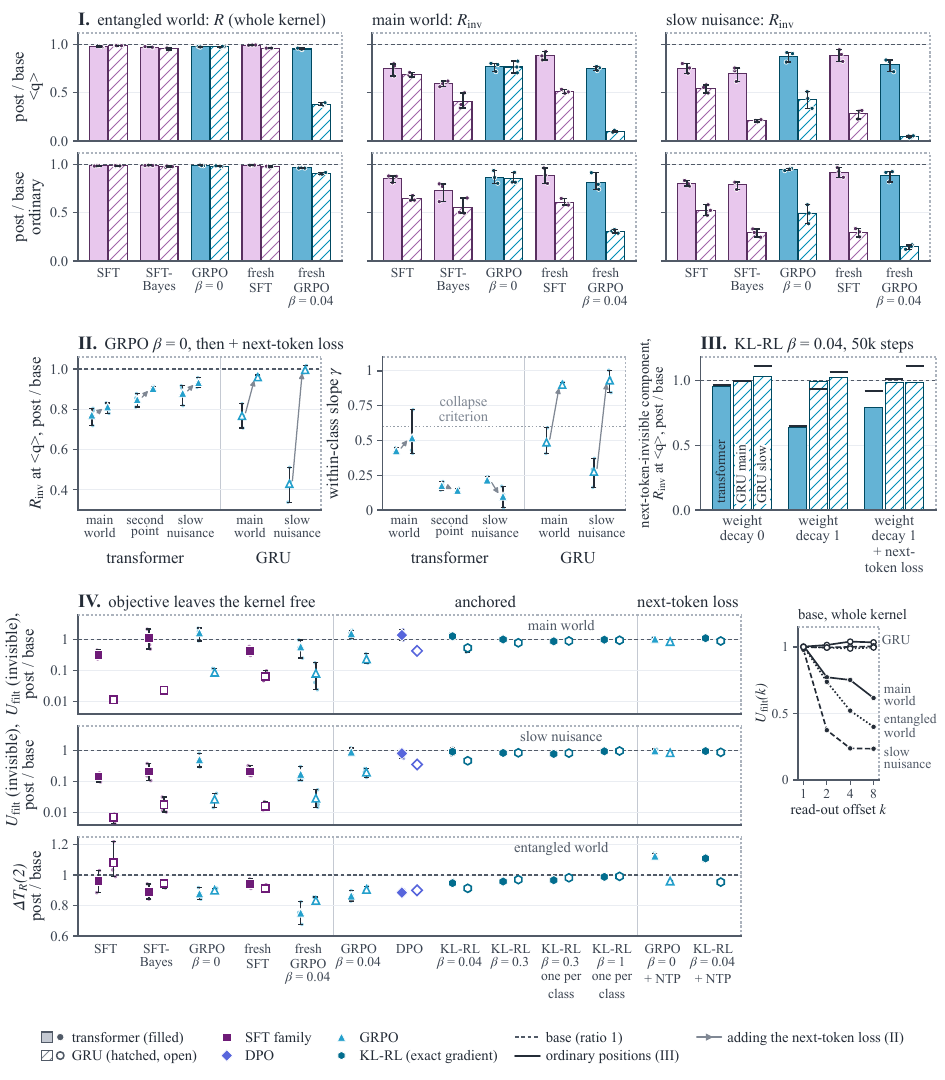}
  \caption{\textbf{Recurrent and attention architectures.} \textbf{I.} Decodability of the free kernel after objectives that leave it free, at the query (top) and at ordinary positions (bottom), for the transformer (filled) and the GRU (hatched): the whole kernel in the entangled world, the next-token-invisible component in two separable worlds. \textbf{II.} Adding a next-token loss to unanchored GRPO (arrows): the invisible component and the within-class slope recover in the GRU and not in the transformer. \textbf{III.} The invisible component under KL-RL at $\beta=0.04$ after 50{,}000 steps, with and without weight decay (lines: ordinary positions). \textbf{IV.} Use of the invisible component by the model's own filter under every objective in two separable worlds, and the change of the reward-relevant statistic two tokens after an exchange in the entangled world. Right: filter use of the whole kernel by the base models over the read-out offset.}
  \label{fig:app-recurrent}
\end{figure*}

\begin{figure*}[t]
  \centering
  \includegraphics[width=\textwidth]{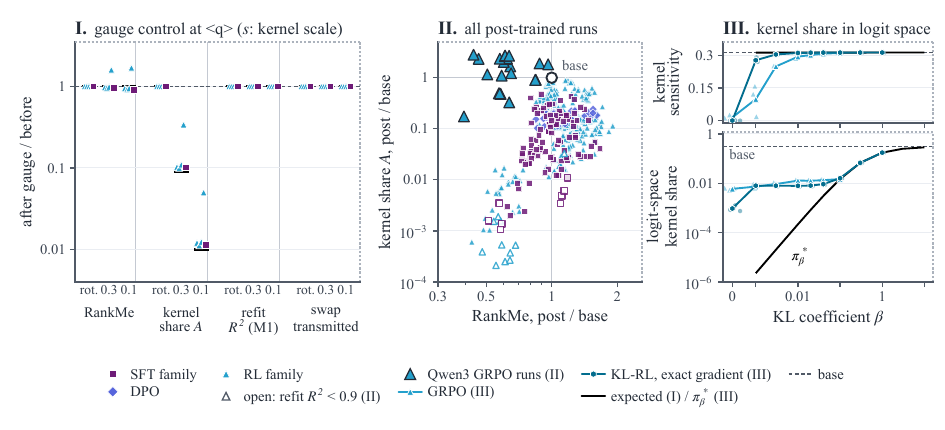}
  \caption{\textbf{Spectral statistics.} The gauge control, the kernel share against the change of effective rank, and the kernel's share of the output variance over $\beta$ against the optimum.}
  \label{fig:app-spectral}
\end{figure*}

\begin{figure}[t]
  \centering
  \includegraphics[width=\columnwidth]{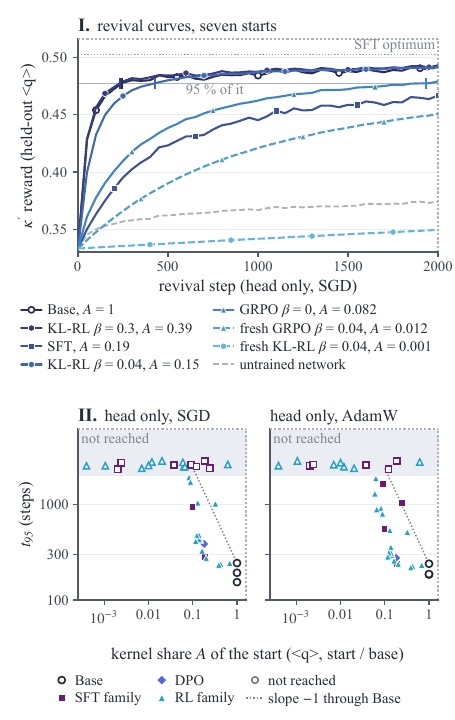}
  \caption{\textbf{Revival.} Reward of a new linear readout for the nuisance variable on frozen networks, and steps to 95\% against the kernel share.}
  \label{fig:app-revival}
\end{figure}

\begin{figure*}[t]
  \centering
  \includegraphics[width=\textwidth]{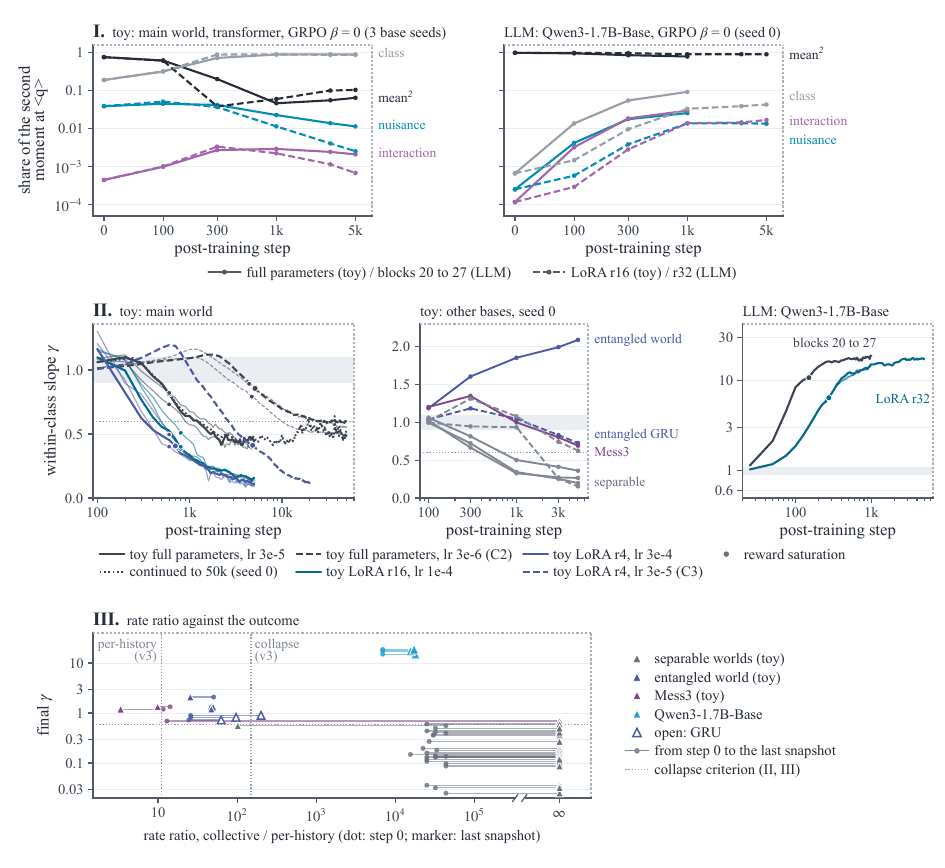}
  \caption{\textbf{How a policy leaves the identity.} \textbf{I.} Shares of the second moment of the query representation carried by its mean, by class features, by nuisance features and by their interaction during unanchored GRPO, in a small transformer (left) and in Qwen3-1.7B-Base (right): the small transformer removes most of the mean within 300 steps, and the language model keeps it. \textbf{II.} Slope $\gamma$ over training across learning rates and parameterizations in the main world, in the other worlds, and in the language model (dots: reward saturation). \textbf{III.} Ratio of the collective to the history-wise rate, from the base model (dot) to the last snapshot (marker), against the final slope.}
  \label{fig:app-direction}
\end{figure*}

\begin{figure*}[p]
  \centering
  \includegraphics[width=\textwidth]{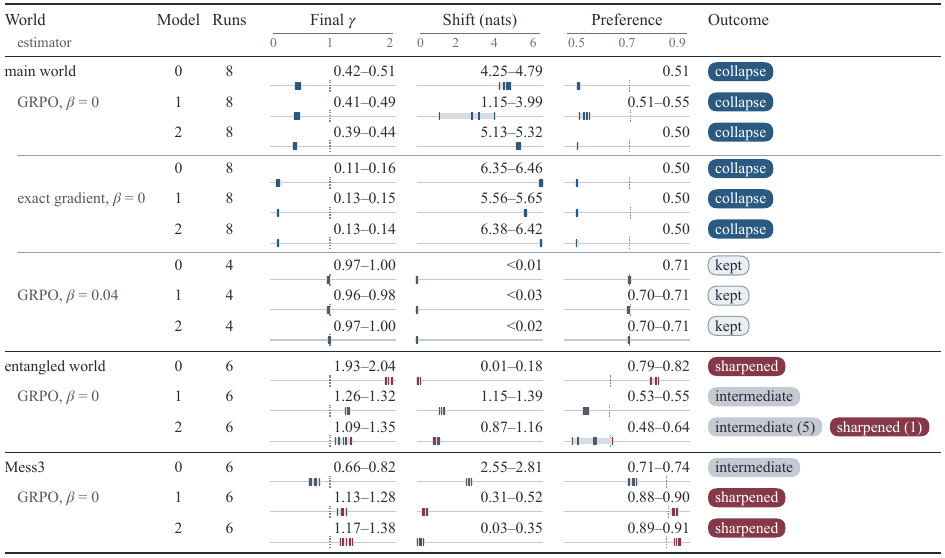}
  \captionof{table}{\textbf{Outcomes over post-training seeds} after 5{,}000 steps. Each cell prints the range over the seeds of one pretrained model and marks every run on a track (dashed: $\gamma=1$, and the base model's preference). Shift: absolute intercept of the policy's within-class log-odds regressed on the reference's. Preference: mean probability, within the favored class, of the token the reference prefers there. Collapse: $\gamma\le0.6$ and preference at most 0.60. Sharpened: $\gamma\ge1.1$ and preference not below the base model's. Kept: $\gamma$ within 0.1 of 1 and shift at most 0.3 nats. Intermediate: none of these.}
  \label{tab:seeds}
  \par\vspace{1.6em}
  \includegraphics[width=\textwidth]{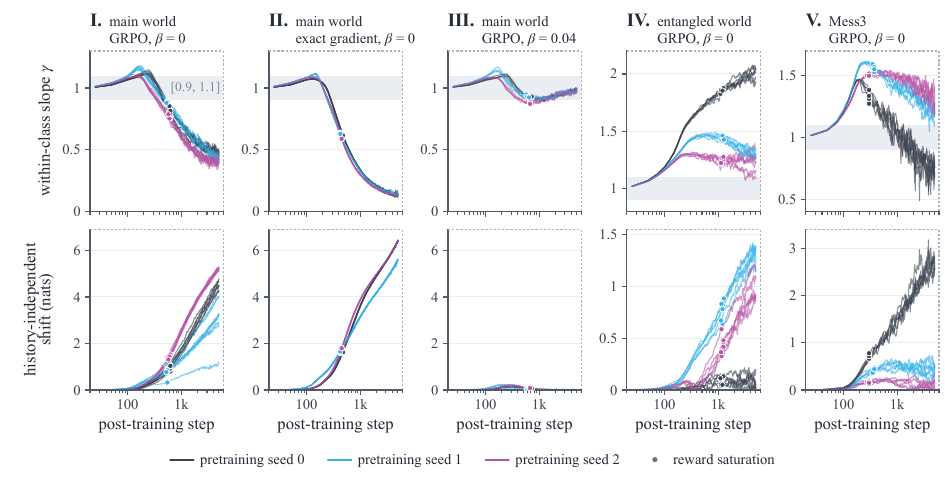}
  \captionof{figure}{\textbf{Every run of the seed study.} Slope $\gamma$ (top) and history-independent shift (bottom) over post-training steps, one line per run, colored by pretrained model, with a dot at reward saturation. Runs from one pretrained model lie on top of each other, and where outcomes differ they differ between pretrained models.}
  \label{fig:app-seeds}
\end{figure*}

\paragraph{Other worlds and architectures.}
Across eight further combinations of world and architecture (Figure~\ref{fig:app-fates}), anchored objectives keep $R\ge0.97$ in all 144 runs, with $\gamma$ between 0.93 and 1.02 for KL-RL and GRPO and between 0.86 and 1.13 for DPO, and the sensitivity of KL-RL policies matches $\pistar$ to within 0.009. SFT reroutes in all 48 runs ($U\le0.26$ except for SFT on Bayes actions in the entangled world, 0.32 to 0.35), with $R\ge0.91$ in transformers and $R\ge0.76$ in GRUs. The fall of the share is a property of the objective at $\beta<1$ and appears in the GRU as well. At $\beta=1$ in the design without ties the share stays at 0.71 to 0.89.

\paragraph{Erosion of the next-token-invisible component.}
Most of the loss of $\Rinv$ occurs within the first 300 steps. Under KL-RL at $\beta=0.04$ the level then holds over 50{,}000 steps (0.95), while unanchored and weakly anchored runs keep losing (from 0.77 to 0.63 without an anchor). An anchored policy at $\beta=0.3$ gives up 60\% of the kernel's share with $R$ within 0.005 of the base, and the effective rank moves in both directions across objectives.

\paragraph{The anchor identity in trained networks.}
At $\beta\ge0.04$ the slope $\gamma$ is between 0.996 and 1.004 for KL-RL and Dr.\ GRPO, between 0.96 and 0.99 for GRPO, and between 0.93 and 1.02 for DPO. Without ties the identity predicts a kernel sensitivity of 0.008, 0.070 and 0.154 at $\beta=0.04$, 0.3 and 1, and the trained policies give 0.009, 0.070 and 0.154. KL-RL at $\beta=1$ and SFT reach rewards of 0.691 and 0.693. Without the anchor the slope falls to 0.43 under GRPO and to 0.12 under the exact gradient, with a history-independent offset of 4.6 to 6.1 nats, while $\Rvis$ stays at 0.96 to 0.97. Figure~\ref{fig:app-anchor} shows the transition between the two regimes over $\beta$.

\paragraph{Two exits.}
Rerouting can be seen within a single network by comparing its two outputs (Figure~\ref{fig:app-exits}). After SFT the policy responds to a twin exchange at 0.15 of the base and the next-token head at 0.56. After KL-RL both stay near 0.8, and after GRPO and DPO near 0.9.

\paragraph{Re-encoding under an anchor.}
Anchored training re-encodes what it keeps. At $\beta=0.04$ the kernel's image in the residual stream rotates by 20 to 77 degrees. An interchange on the subspace fitted at the base model then transfers only 0.78 to 0.88 of the twin effect and a probe frozen at the base model fails, where the subspace and the probe refit on the post-trained model show the information and its use intact (0.94 to 0.97). A frozen diagnostic would read re-encoding as erasure, which is why every coordinate here is refit on the model under test.

\paragraph{Matched KL.}
The divergence at the policy level is not a consequence of unanchored runs travelling farther. At the KL from the reference that anchored runs reach for $\beta$ between 0.01 and 0.1 (0.38), unanchored AdamW runs have the same $\Rinv$ (0.86 to 0.87 against 0.82 to 0.94) and a within-class preference already below every anchored run (Figures~\ref{fig:app-anchor} and~\ref{fig:app-matched}).

\paragraph{The weight-decay stress test.}
The erasure is late: under SFT $\Rinv$ is 0.69, 0.30 and 0.001 after 10, 30 and 50 thousand steps (Figure~\ref{fig:app-erasure}). In the anchored policy an added next-token loss raises the retained 0.64 of the invisible component to 0.79.

\paragraph{The recurrent model.}
Under objectives that free the kernel the GRU retains less of the next-token-invisible component than the transformer at the query position, by 0.19 to 0.75 for SFT on Bayes actions and for the fresh-action design, and it loses the component at ordinary positions as well (0.14 to 0.53 against 0.79 to 0.95 in the world with a slow nuisance chain). A next-token loss restores the GRU almost completely (0.96 to 0.99) and turns the collapse without an anchor into an intermediate outcome in six of six GRU runs, while all twelve transformer runs with the same loss in the separable worlds still collapse: the next-token term at ordinary positions constrains the query through the recurrent state and cannot reach it through attention. In the weight-decay corner the GRU erases along the same line as the transformer. The invisible component falls to at most 0.03 under SFT and the fresh-action objectives, at ordinary positions as well and on the same timescale (Figures~\ref{fig:app-erasure} and~\ref{fig:app-recurrent}), and the filter use of the erased runs is $10^{-5}$ to $10^{-3}$ of the base. Figure~\ref{fig:recurrent} sets use against representation: after objectives that free the kernel the GRU's filter uses the invisible component at 0.01 to 0.09 of the base while 0.15 to 0.85 of it is still decodable at ordinary positions (0.53 to 0.65 after SFT), and a next-token loss restores the use to 0.83 to 0.88. A transformer recomputes part of an exchanged component from the tokens left in place (0.37 to 0.77 of the Bayes effect two tokens later at the base model), so its ratios are noisier. Under KL-RL at $\beta=0.04$ the filter keeps 0.46 to 0.53 of its use of the invisible component and 0.8 of its use of the whole kernel, and across anchor strengths the retained use ranges from a fifth to four fifths. The base GRU's filter uses the kernel as the exact Bayes filter does (ratio 0.99 to 1.04 at every lag). In the entangled world, where the kernel feeds future reward-relevant beliefs, the filter keeps using it under every objective (0.75 to 1.13).

\paragraph{The invisible component in the filter-use test.}
Exchanging the invisible component alone gives $\Ufilt$ of 0.40 to 0.51 for the base GRU two tokens later. A belief with only this component exchanged lies off the set of reachable nuisance beliefs, and the ratio falls with the distance from that set. Exchanging the whole kernel maps a reachable belief to a reachable belief and gives $\Ufilt\approx1$ at every lag, which the invisible component could not allow if the filter ignored it, since it is invisible one token later by construction.

\paragraph{Gauge control.}
Scaling the kernel's image in a trained network by 0.1 and compensating at the readout moves the kernel's share twenty- to a hundredfold and the effective rank by a factor of 0.9 to 1.7, with the probe $R^{2}$ and the interchange effect unchanged to numerical precision (Figure~\ref{fig:app-spectral}).

\paragraph{Revival.}
A new linear readout for the nuisance variable reaches 95\% of the attainable reward in 154 to 244 steps on the base model, 233 to 432 after anchored objectives, 943 to 1{,}932 or not within 2{,}000 steps after unanchored and fresh-action objectives, and not within 2{,}000 steps after SFT or wherever the kernel share is below about one hundredth of the base (Figure~\ref{fig:app-revival}). Two kinds of starting point are slower than an untrained network even under full fine-tuning: the weight-decay corner, where a readout on frozen features stays at chance and full fine-tuning needs 934 to 1{,}143 steps against 429 from scratch, and by a smaller margin the fresh-action KL-RL policy (488 steps). The same corner with an added next-token loss recovers faster (832 steps) and remains 2.4 times slower than the base model.

\subsection{The direction without the anchor}
\label{app:direction}

The collapse does not depend on sampling noise, on the tying of the head, or on the optimizer family. The exact expected gradient collapses more completely than GRPO (intercept 6.1 nats). An untied head with a trainable trunk collapses like the tied head, with the same slopes within seed noise, and the collapsed log-odds are written in the residual stream. Training only the head does not collapse within 5{,}000 steps, and its trajectory is predicted without free parameters by the fixed-feature model of Appendix~\ref{app:dynamics}. SGD collapses more slowly than AdamW. A next-token loss does not prevent the collapse in transformers. The setting matters in practice because current recipes for reinforcement learning with verifiable rewards mostly remove the KL term \citep{yu2025,liu2025,hu2025}, after earlier ones used coefficients of 0.04 and 0.001 \citep{shao2024,guo2025}, which is the range in which the policy-level threshold of Section~\ref{sec:anchor} lies.

Parameterization and budget do not change the direction (Figure~\ref{fig:app-direction}). Low-rank adapters of rank 4 and 16 collapse in nine of nine runs and remove the mean of the query representation faster than full fine-tuning. At a tenth of the learning rate the reward saturates eight times later and the slope still falls to between 0.50 and 0.60. The slope first rises above 1 and falls below 1 before the reward saturates (at the highest adapter learning rate the first record already lies past the peak).

In worlds where the two rates are comparable at the base model (a ratio between 10 and 50 in the entangled world and in Mess3, against $10^{4}$ in the separable worlds), the first 300 steps sharpen and the outcome after 5{,}000 steps ranges from sharpening to a partial collapse.

\paragraph{Post-training seeds.}
The runs above use one post-training seed per pretrained model. Repeating the unanchored runs of the main world with eight further seeds for each of three pretrained models gives the same outcome 48 times (Table~\ref{tab:seeds}, Figure~\ref{fig:app-seeds}). Under GRPO the slope peaks between 1.09 and 1.18 after 125 to 250 steps, falls below 1 between steps 250 and 425, and the reward saturates between steps 525 and 625. Under the exact gradient the sequence is the same and faster. The twelve anchored controls ($\beta=0.04$) return to a slope between 0.96 and 1.00.

The token a class collapses onto is fixed before post-training starts. In 17 of the 18 combinations of pretrained model, class and estimator all eight seeds select the same token, GRPO and the exact gradient agree in all nine combinations of model and class, and within a model the three classes select the same nuisance symbol, the combination that Appendix~\ref{app:dynamics} identifies as the fastest-growing one. The selected token is unrelated to the reference's average preference between the two, which it matches in five of the nine combinations of model and class. The exception shows where sampling enters. In one class of one pretrained model the collective mode is slow under GRPO (the winning token holds between 0.50 and 0.95 of the class after 5{,}000 steps), two of the eight seeds select the other token, and in one of them the three classes shift in opposite directions ($-2.5$, $4.7$ and $1.4$ nats), which leaves a pooled shift of 1.15 nats, the low end of the range in Table~\ref{tab:seeds}. The exact gradient selects the same token eight times in this class as well.

In the entangled world and in Mess3 the seeds of one pretrained model again end together, and the pretrained models end apart. Within a model the final slope varies by at most 0.26 across seeds. Between models it ranges from 1.09 to 2.04 in the entangled world and from 0.66 to 1.38 in Mess3. In the entangled world one model sharpens and two shift by about a nat with the slope still above 1. In Mess3 two models sharpen and one shifts by 2.6 to 2.8 nats with the slope below 1.

\section{Language-Model Experiments}
\label{app:llm}

\begin{figure*}[t]
  \centering
  \includegraphics[width=\textwidth]{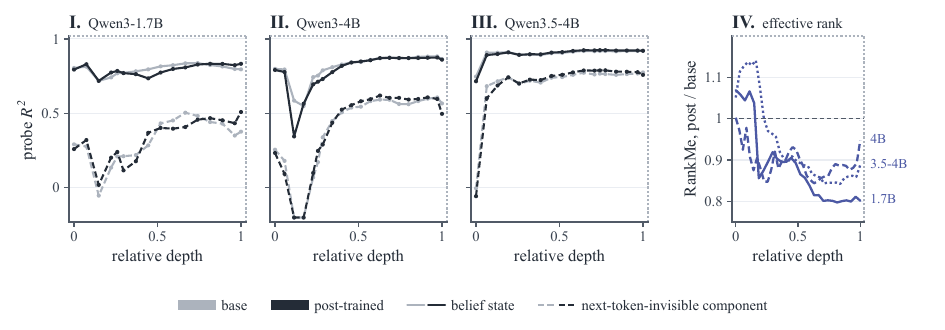}
  \caption{\textbf{Released checkpoints.} Decodability of the in-context belief state and of its next-token-invisible component by layer for base and post-trained checkpoints (means over six hidden Markov models), with the effective rank of the same layers.}
  \label{fig:app-llm}
\end{figure*}

\paragraph{Released checkpoints.}
We follow the protocol and the released code of \citet{balcells2026}: a stream of 20{,}000 tokens from one of six hidden Markov models, linear probes on the residual stream of every layer, and steering along probed directions. Steering moves the predictions after post-training as before. Figure~\ref{fig:app-llm} shows the probes by layer, averaged over the six processes, while the peak values quoted in the text are taken per process. The invisible component is scored with probes trained on other sequences and tested on a held-out one, after truncated principal components and ridge, because interleaved splits within one sequence reward slowly mixing processes. Over the pairs the share of variance carried by the invisible component rises by 9 to 40\%. Peak $R^{2}$ of the belief state is between 0.85 and 0.99 in the Qwen pairs, within the range reported for six models by \citet{balcells2026}. A base model of another family did not converge to the Bayes predictor in context and was excluded before any comparison.

\paragraph{Controlled GRPO.}
The stream uses six letters that are single tokens in context, two per class, and a marker that is followed by the next observation, inserted with probability 0.2. Decisions are the markers in the second half of a context of 2{,}048 tokens, about 170 per prompt. The policy is the next-token distribution restricted to the six letters and the reference is the same model with the adapters switched off. Adapters have rank 32 on all linear maps of every block, the optimizer is AdamW at $10^{-6}$ with four prompts per step, and runs last 1{,}000 steps unless stated. Measurements use 128 twin pairs with 43{,}582 decisions. At the base model the slope of the model's within-class log-odds against the Bayes log-odds is 0.63 to 0.72. With $\beta\ge0.3$ the slope against the reference returns to between 1.00 and 1.08 within 1{,}000 steps. With $\beta=0.04$ it overshoots six- to ninefold and, in the run we continued, returns between 4{,}000 and 10{,}000 steps. Both kernel components stay decodable in every run ($R\ge1.0$). The kernel's share rises under amplification (1.1 to 2.8 times the base) and falls under an anchor of $\beta=0.3$ (0.33 to 0.49).

\paragraph{Controls.}
The exact gradient and Dr.\ GRPO amplify less than GRPO at $\beta=0.04$ (3.2 and 4.5 against 9.4) and in the same direction. A next-token loss restores ordinary positions and leaves the amplification at the marker in place. Fully fine-tuning the last eight blocks amplifies faster than adapters and to a similar level (19.9 against 16.0). Continuing the unanchored run to 5{,}000 steps, seventeen times the time the reward needs to saturate, raises the slope from 15.0 to 17.4. In all unanchored runs the mean of the representation at the marker stays in place (0.80 to 0.95 of the second moment, against 0.997 at the base model) while its class component grows 31 to 110 times, whereas the small transformers remove most of the mean within 300 steps. The effective rank falls by 16 to 61\% in the 1.7B model.

\end{document}